\documentclass[11pt]{article}

\usepackage[utf8]{inputenc}
\usepackage[T1]{fontenc}
\usepackage{lmodern}
\usepackage[letterpaper,margin=1in]{geometry}
\usepackage{microtype}

\usepackage{amsmath,amssymb}
\usepackage{graphicx}
\usepackage{array}            
\usepackage{longtable}        
\usepackage{booktabs}         
\usepackage{tabularx}         
\usepackage{xltabular}        
\usepackage{ragged2e}         
\usepackage[normalem]{ulem}

\newcolumntype{Y}[1]{>{\hsize=#1\hsize\RaggedRight\arraybackslash\hyphenpenalty=2000}X}
\newcolumntype{Z}{>{\RaggedRight\arraybackslash\hyphenpenalty=2000}X}

\usepackage[font=footnotesize,labelfont=bf,labelsep=period,justification=justified,
            singlelinecheck=false,skip=6pt]{caption}
\usepackage[round,authoryear]{natbib}
\usepackage{xcolor}
\definecolor{citeblue}{RGB}{0,84,166}
\newcommand{\rev}[1]{#1}
\usepackage[colorlinks=true,citecolor=citeblue,linkcolor=citeblue,urlcolor=citeblue,
            breaklinks=true]{hyperref}   

\providecommand{\doi}{}
\renewcommand{\doi}[1]{\href{https://doi.org/#1}{doi:\,\begingroup\urlstyle{rm}\nolinkurl{#1}\endgroup}}

\graphicspath{{images/}}

\newcommand{\xref}[3][]{\textbf{\hyperref[#3]{#2~\ref*{#3}#1}}}
\newcommand{\nref}[1]{\textbf{\hyperref[#1]{\ref*{#1}}}}

\newcommand{\tabfont}{\footnotesize}

\newcommand{\notefont}{\scriptsize}

\bibpunct{(}{)}{;}{a}{,}{,}   

\title{PlacePulse: Learning Transferable Place Representations from
Structural Geographic Data}

\author{
  Luc\'{\i}a Garc\'{\i}a-Duarte\thanks{Corresponding authors:
  \texttt{lgarciaduarte@carto.com}, \texttt{malvarez@carto.com}},
  Miguel \'Alvarez\footnotemark[1], \\
  Diego Vicente,
  Javier P\'erez Trufero,
  Javier de la Torre \\[0.5em]
  (CARTO)
}

\date{}

\begin{document}
\maketitle

\begin{abstract}
Geospatial machine learning often depends on task-specific feature engineering over large collections of sociodemographic, business, and environmental variables. Geospatial foundation models offer reusable representations of place, but leading approaches typically encode dynamic signals, such as mobility and search activity, or visual signals from satellite imagery. Few models represent the static structural composition of places (who lives and works there, which businesses operate there, and which risks they face) as transferable embeddings at fine spatial resolution. We introduce PlacePulse, a structural geospatial foundation model that compresses 572 attributes, expanded to 2,288 inputs through H3 neighborhood aggregates, into 256-dimensional unit-norm embeddings for 1,055,537 H3 resolution-7 cells in the United States. We also introduce a reusable two-axis benchmark that separates intrinsic representation quality from downstream predictive utility across multiple targets and spatial supports. Intrinsically, the embeddings reconstruct standardized inputs with mean $R^2 = 0.853$, use 251 of 256 approximately decorrelated dimensions, and show strong spatial organization, with Moran's $I = 0.93$ on the leading principal component and autocorrelation persisting beyond the input radius. Extrinsically, across 74 held-out targets from 12 external sources and four spatial supports, PlacePulse outperforms a dimensionality-matched feature-selection baseline on targets unrelated to the training inputs and is competitive with Google's Population Dynamics Foundation Model, leading on health, economic, election, economic-mobility and social-vulnerability outcomes, with relative strengths that vary by domain. These findings position structural composition as a distinct and complementary modality for geospatial foundation models, not simply a fallback where behavioral or imagery data are unavailable.
\end{abstract}



\section{Introduction}
\label{sec:intro}

Geographic analysis increasingly depends on the ability to characterize places \citep{singleton2026}. Whether the objective is to identify underserved neighborhoods, estimate environmental risk, predict retail demand, or support public policy, analysts typically rely on large collections of demographic, socioeconomic, business, and environmental variables to describe the composition of geographic areas \citep{elbers2003, luo2003, mitchell2015, roigtierno2013, singleton2014}. In practice, however, constructing these representations remains a labor-intensive process. Even relatively simple analyses require assembling data from multiple sources, harmonizing spatial units, engineering derived variables, and selecting task-specific features \citep{noardo2022}. As the volume and diversity of available geospatial data continue to grow, this repeated feature engineering has become one of the principal bottlenecks in applied spatial modeling \citep{janowicz2020, li2016, singleton2026}.

Foundation models offer an alternative paradigm. Rather than designing features separately for each application, a model is pretrained once on broad data to learn transferable representations that can be reused across many downstream tasks with limited additional supervision \citep{bommasani2021}. In the geospatial domain, this idea has evolved from early region representation learning, where neighborhoods or spatial units were encoded as dense embeddings \citep{yao2018}, into a rapidly expanding family of geospatial foundation models capable of producing general-purpose representations at increasingly large spatial scales \citep{zhang2024, chen2025}. The central objective is no longer to engineer variables for individual analyses, but to learn reusable representations that capture meaningful characteristics of place.

Although the field has expanded rapidly, current geospatial foundation models have largely converged around two dominant sources of information. One family learns representations from dynamic behavioral signals, including mobility trajectories, visitation patterns, web-search activity, and related traces of human activity. The Population Dynamics Foundation Model (PDFM) \citep{agarwal2024} exemplifies this approach, demonstrating that behavioral signals can support transferable geographic representations across diverse downstream prediction tasks. A second family derives representations directly from visual observations, using self-supervised learning over satellite or street-level imagery to encode the physical appearance of locations \citep{cong2022, yan2024, brown2025}. These two paradigms have substantially advanced geographic representation learning and established the current foundation-model landscape.

Yet many applications in geography require representations that capture characteristics of places that are comparatively stable over time and not directly observable from either behavioral traces or visual appearance. We refer to this family of information as structural information: the demographic, socioeconomic, commercial, functional, and environmental attributes that describe how places are organized. Such information underpins a wide range of applications in urban planning, public health, environmental assessment, market analysis, and policy evaluation \citep{elbers2003, elliott2004, mitchell2015, singleton2014, wangyin2023}. Despite its central role in geographic analysis, structural information has received comparatively little attention as the primary basis for learning transferable geographic representations. Existing work has typically incorporated structural variables as auxiliary inputs within multimodal models or combined a limited set of demographic attributes with mobility signals in city-scale settings \citep{cao2025, huang2022, wang2025calliper, wen2024}. This leaves open the broader question of whether structural information can itself support general-purpose geographic representations that transfer across downstream tasks.

This gap is not merely methodological; it is also conceptual. Current foundation models implicitly emphasize two complementary perspectives on geographic space: behavioral representations encode how places are used, whereas imagery-based representations encode what places look like. We argue that a third perspective deserves independent investigation: structural representations, which encode what places are through their relatively stable socioeconomic, commercial, functional, and environmental composition. Rather than viewing structural variables simply as covariates to be added to behavioral or imagery models, we investigate whether they constitute a distinct representation modality capable of supporting transferable geographic embeddings in their own right.

To address this question, we develop a geospatial foundation model pretrained exclusively on structural attributes derived from the Applied Geographic Solutions (AGS) national database \citep{ags2025} (\xref{Figure}{fig:overview}). The model is trained in a self-supervised manner using a masked autoencoder objective \citep{he2022} over 572 structural variables, augmented with features derived from neighboring cells to capture local spatial context. It produces compact 256-dimensional vector embeddings for more than one million H3 resolution-7 cells \citep{uber2018} covering the United States. The embeddings are constrained to the unit hypersphere, so cosine similarity is equivalent to the dot product. Nearest-neighbor retrieval across the full million-cell grid can therefore be implemented as a single matrix product, enabling interactive similarity search at national scale. We refer to the resulting representations as PlacePulse embeddings.

\begin{figure}[tbp]
\centering
\includegraphics[width=\linewidth]{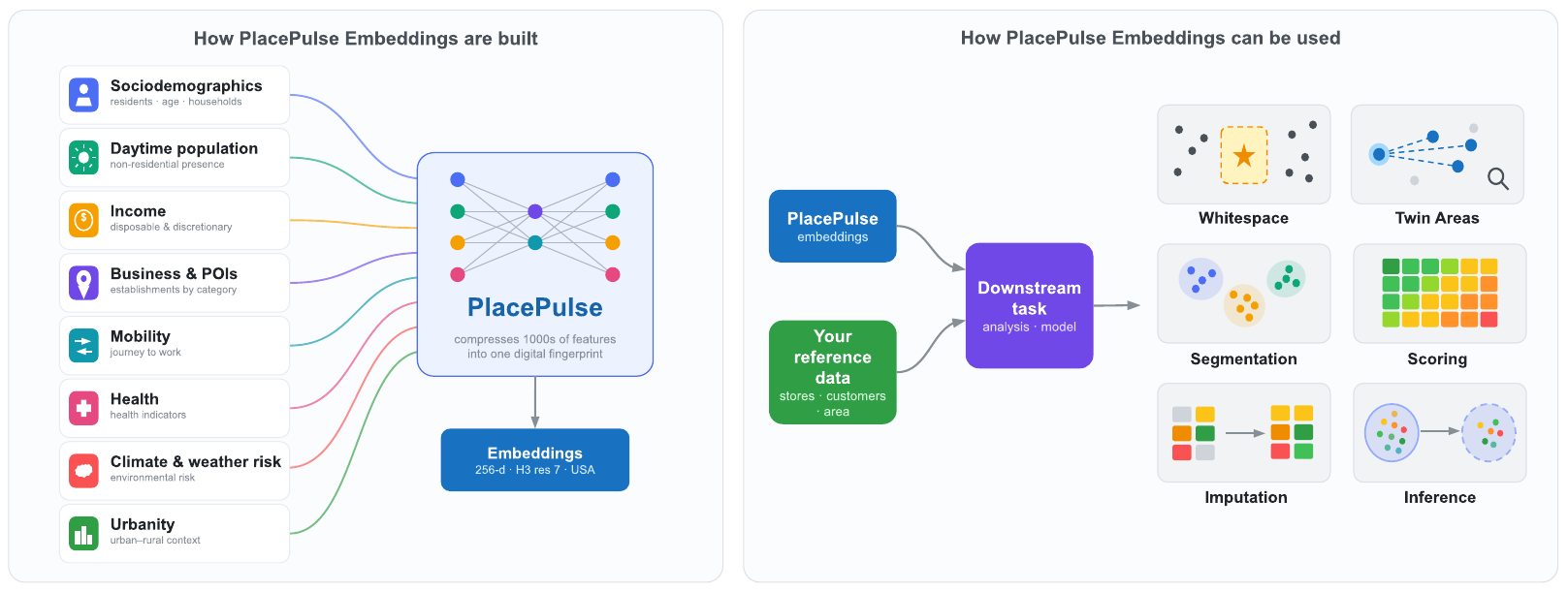}
\caption{\textbf{Overview of the geospatial foundation model: how the embeddings are extracted and applied to downstream tasks.}
A self-supervised autoencoder compresses a broad set of structural attributes per location (sociodemographic, economic, health, climate, mobility, and others) into a single 256-dimensional, L2-normalized embedding at H3 resolution 7. This representation drives two families of applications: (1)~similarity-based retrieval for tasks such as white-space analysis, twin-area and look-alike discovery, cannibalization and site screening, or market segmentation; and (2)~supervised prediction, where the embeddings serve as off-the-shelf features for regression and classification at the native H3 resolution or aggregated to the target's spatial unit.}
\label{fig:overview}
\end{figure}

PlacePulse embeddings are evaluated from two complementary perspectives. Internally, we assess whether the embeddings preserve the structural information from which they are learned and examine the organization of the resulting latent space. Externally, we evaluate whether the representations transfer to downstream prediction tasks across different targets and spatial supports, including comparisons with PDFM. This allows us to distinguish the intrinsic properties of the learned representation from its utility as a transferable feature set.

Our contributions are:

\begin{enumerate}
\item \textbf{A new foundation model built on the structural characteristics of place.} We develop PlacePulse, a self-supervised model that learns transferable geographic representations from demographic, socioeconomic, commercial, functional, and environmental attributes. PlacePulse compresses 572 attributes and their neighborhood aggregates into 256-dimensional embeddings for more than one million H3 resolution-7 cells across the United States (\xref{Figure}{fig:pca}).

\item \textbf{A comprehensive benchmark of representation quality and predictive transfer.} We introduce a two-axis evaluation framework that combines intrinsic assessment of reconstruction fidelity, latent-space structure, effective dimensionality, and spatial organization with extrinsic evaluation across 74 targets from 12 external sources and four spatial supports, using a fixed estimator and state-blocked cross-validation. Because the framework is independent of the representation architecture and input data, it provides a common basis for comparing different geospatial embeddings.

\item \textbf{A systematic and transparent empirical evaluation of structural embeddings.} We apply the proposed benchmark to characterize PlacePulse across intrinsic properties, downstream tasks, and spatial supports, reporting both its strengths and limitations. PlacePulse produces a compact and spatially coherent representation and outperforms a dimensionality-matched, target-blind baseline on targets not used to train the autoencoder. Comparison with PDFM reveals that neither representation consistently outperforms the other across domains and spatial supports, suggesting that structural and behavioral representations capture different aspects of place.

\end{enumerate}

\begin{figure}[tbp]
\centering
\includegraphics[width=\linewidth]{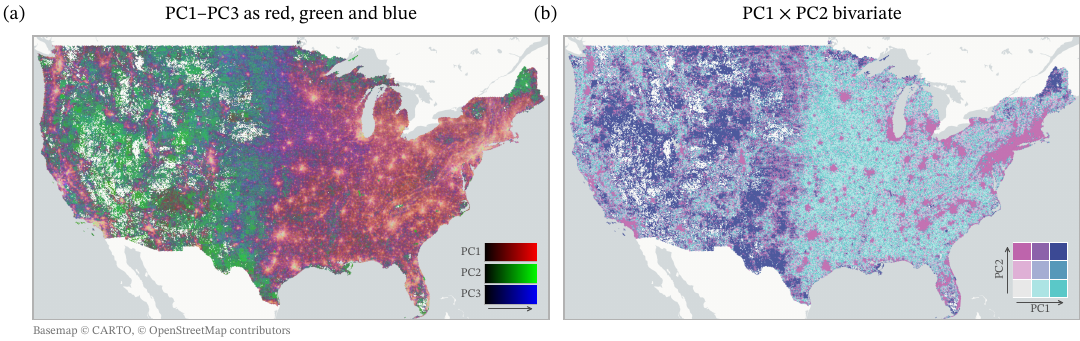}
\caption{\textbf{Visualizations of the presented embeddings across the contiguous United States (H3 resolution 7, ${\sim}$1.06M cells), colored by principal component analysis (PCA) of the 256-dimensional embeddings.}
(a) The three leading components (PC1--PC3) are min--max scaled to the red, green, and blue channels of an image; cells with similar colors have similar embeddings. (b) A bivariate map of the two leading components (PC1 and PC2), each split into terciles and combined into a $3\times3$ color scheme, from low--low to high--high.}
\label{fig:pca}
\end{figure}

The remainder of the paper is organized as follows.
\xref{Section}{sec:related} reviews related work on geospatial foundation models and region representation learning and positions our contribution.
\xref{Section}{sec:data} describes the AGS data and preprocessing.
\xref{Section}{sec:methods} details the model architecture, training, and evaluation protocol.
\xref{Section}{sec:results} reports internal and downstream results.
\xref{Sections}{sec:discussion} and~\nref{sec:conclusion} discuss implications, limitations, and conclusions.

\section{Related work}
\label{sec:related}

This section reviews the evolution of geographic representation learning from early region embeddings to contemporary geospatial foundation models, summarizes the principal self-supervised learning paradigms underlying these models, and positions structural representations of place within this broader landscape.

\subsection{From region embeddings to geospatial foundation models}
\label{sec:rw-region}

The idea that geographic regions can be represented as dense vectors emerged in urban computing, where embeddings were learned to summarize neighborhoods or functional zones and transferred across downstream prediction tasks \citep{yao2018}. Rather than relying on manually engineered geographic indicators, these methods sought to learn latent representations directly from heterogeneous urban data \citep{wang2020urban2vec, huang2022}.

One of the earliest examples is the work of \citet{yao2018}, who embedded urban zones from human mobility patterns to recover functional similarity between neighborhoods. This established the now common view of a geographic region as a learnable representation rather than merely a collection of attributes. Subsequent work progressively incorporated additional data modalities. Urban2Vec combined street-view imagery with points of interest to produce multimodal neighborhood embeddings \citep{wang2020urban2vec}; \citet{huang2022} learned semantic representations from POI distributions; UrbanCLIP extended CLIP-style language-image pretraining to urban regions \citep{yan2024}; and CaLLiPer employed multimodal contrastive learning over POIs to learn transferable urban representations while explicitly framing the resulting embeddings as geospatial foundation models \citep{wang2025calliper}.

Across this literature, three characteristics are remarkably consistent. First, embeddings are intended to be reusable representations rather than task-specific features. Second, information from multiple modalities is increasingly integrated to characterize place. Third, the spatial domain is typically limited to one or a small number of metropolitan areas. These developments established the conceptual foundations from which today's geospatial foundation models emerged.

\subsection{Geospatial foundation models}
\label{sec:rw-gfm}

Inspired by the broader foundation-model paradigm \citep{bommasani2021}, recent work has shifted from city-specific embeddings toward large-scale pretrained representations intended for general downstream use. Recent surveys by \citet{zhang2024}, \citet{chen2025}, and \citet{zhou2026} document the rapid development of geospatial representation learning toward increasingly general-purpose and foundation-model approaches, with substantial attention to the role of different data modalities and heterogeneous spatial context, particularly remote sensing imagery and other heterogeneous geospatial observations.

The first major family learns representations from dynamic human-behavior signals, including mobility trajectories, visitation patterns, web-search activity, and related behavioral traces. PDFM is the canonical example. It combines cartographic information, mobility-derived busyness, search trends, and environmental variables within a graph neural network to produce transferable embeddings for U.S. ZIP Code Tabulation Areas and counties. Its demonstrated performance across diverse socioeconomic, health, and retail prediction tasks illustrates the effectiveness of behavioral signals as general-purpose geographic representations.

The second dominant family derives representations directly from Earth-observation and visual imagery. Early work adapted masked-image modeling to remote sensing through models such as SatMAE \citep{cong2022}, Scale-MAE \citep{reed2023}, Prithvi \citep{jakubik2023}, and Clay \citep{clay2024}. Parallel efforts have employed contrastive objectives, including SatCLIP \citep{klemmer2025}, which aligns satellite imagery with geographic coordinates, and UrbanCLIP \citep{yan2024}, which aligns imagery with textual descriptions. More recently, global embedding-field models such as AlphaEarth Foundations \citep{brown2025} and TESSERA \citep{feng2025} produce dense embeddings for virtually every terrestrial location at fine spatial resolution.

Although these two families differ substantially in their data sources, their representations are primarily shaped either by how places are used (behavioral traces) or by what places look like (imagery). Structural information, though incorporated as auxiliary information, has received comparatively less attention as a primary signal for learning general-purpose geographic representations. This motivates investigating whether structural information alone can support transferable representations of place.

\subsection{Self-supervised learning for geospatial representations}
\label{sec:rw-ssl}

The success of modern geospatial representations is closely tied to advances in self-supervised learning, where representations are learned without downstream task labels by optimizing objectives defined directly on the input data.

Three broad paradigms dominate current geospatial self-supervised learning. Reconstruction-based methods learn compressed representations by recovering masked or corrupted inputs, an approach widely adopted in remote sensing \citep{cong2022, reed2023, jakubik2023}. Contrastive methods instead maximize agreement between different views of the same location while separating unrelated locations, forming the basis of Tile2Vec \citep{jean2019}, SatCLIP \citep{klemmer2025}, UrbanCLIP \citep{yan2024}, and CaLLiPer \citep{wang2025calliper}. A third family learns representations by predicting spatial or temporal context, particularly in trajectory- and graph-based models \citep{lin2021}.

Our model belongs to the reconstruction-based family. Like previous self-supervised approaches, it learns a transferable latent representation without downstream supervision. Unlike most existing work, however, the reconstruction objective operates on high-dimensional structural tabular data rather than imagery, trajectories, or graph neighborhoods. As noted by \citet{chen2025}, structural tabular attributes remain comparatively underexplored as a pretraining signal despite their importance for many applied geographic analyses.

\subsection{Structural representations of place}
\label{sec:rw-structural}

While most geospatial foundation models derive representations from behavioral traces or imagery, many applied spatial analyses depend primarily on structural descriptions of place. We use the term structural representation to refer to a representation learned from relatively stable attributes that characterize the composition and functional context of a location, rather than its observed activity patterns or visual appearance. These attributes may include sociodemographic characteristics, income and employment structure, business composition, commuting patterns, health conditions, built-environment indicators, and exposure to environmental and climate-related risks.

Despite the central role of structural attributes in applied geography, urban planning, public health, environmental assessment, and market analysis, they have received comparatively little attention as the primary signal for transferable representation learning. Existing approaches generally use such information in one of three ways. First, demographic, socioeconomic, or land-use variables may appear as auxiliary inputs within multimodal models dominated by mobility, imagery, or textual data \citep{cao2025}. Second, POI- and semantics-based approaches capture specific components of urban structure, particularly land use, business activity, and urban function, without representing the broader composition of place \citep{huang2022, wang2025calliper}. Third, a smaller body of work explicitly treats demographic composition as a learnable regional representation \citep{wen2024}.

The closest work to ours is Demo2Vec \citep{wen2024}, which learns region embeddings using demographic information together with human mobility. Demo2Vec shares our premise that demographic composition can form part of a transferable representation rather than serving only as a set of task-specific covariates. It therefore provides an important demonstration that population structure contains reusable geographic information. However, it is evaluated at the scale of New York and Chicago, combines demographic variables with mobility traces, and focuses on a narrower conception of structural information than the one considered here.

Our notion of structural representation is substantially broader than previous work. Rather than encoding a single component of place, such as demographics, POIs, or land use, we seek to represent its overall structural composition through a wide range of sociodemographic, economic, commercial, commuting, health, and environmental attributes. We also isolate this modality by excluding behavioral and imagery inputs, allowing its independent contribution to downstream prediction to be evaluated directly.

\subsection{Positioning our work}
\label{sec:rw-positioning}

The literature reviewed above suggests that geospatial foundation models have largely converged on two dominant sources of information: behavioral signals and visual observations. Structural information, despite underpinning many applied geographic analyses, has mostly remained an auxiliary source of information or has been explored only in city-scale settings.

Our work investigates whether structural information can itself support a general-purpose geospatial foundation model. We pretrain exclusively on a broad collection of structural attributes describing the socioeconomic, commercial, functional, and environmental composition of places, producing transferable embeddings at a fine national scale. Rather than positioning structural representations as a replacement for behavioral or imagery-based models, we investigate them as a distinct representation modality and compare them directly with PDFM. By evaluating both representations across the same downstream tasks, we examine whether their transfer performance reveals differences in the information captured by structural and behavioral signals. We leave explicit fusion of these modalities to future work.

\section{Data}
\label{sec:data}

The proposed foundation model is trained on a comprehensive collection of harmonized structural attributes describing the demographic, socioeconomic, commercial, functional, and environmental characteristics of places across the United States. These attributes are assembled from multiple AGS products and provide a consistent representation of each H3 spatial cell at national scale. This section covers the data source and spatial frame (\xref{Section}{sec:frame}), the thematic feature families that form the base attribute set (\xref{Section}{sec:inputs}), the ring-aggregate features encoding neighborhood context (\xref{Section}{sec:ring}), and the external data used only as downstream evaluation targets (\xref{Section}{sec:targets}).

\subsection{Source and spatial frame}
\label{sec:frame}

We build the model on the \textit{2025 Applied Geographic Solutions (AGS) demographic, consumer, business, and environmental databases} for the United States, indexed on Uber's H3 hierarchical hexagonal grid at resolution 7 ($\approx$5~km\textsuperscript{2} per cell). The grid covers the populated U.S. footprint at 1,055,537 cells used for training and embedding generation. 

We use the vendor-provided H3 resolution-7 product as the native analytical support, avoiding an additional disaggregation or areal-interpolation step before training. H3 provides a consistent hierarchical index and efficient neighborhood operations, although cell areas are not perfectly constant and the choice of resolution still defines the spatial support of the learned representation.

The AGS estimates are modeled current-year values rather than direct observations: H3 res-7 level demographics are derived from private and public sources, including the 2020 Census and the American Community Survey. Full details of the estimation methodology and source data are documented by the vendor \citep{ags2025b}.

\subsection{Input features}
\label{sec:inputs}

The base feature set is assembled from several AGS source tables at H3 resolution 7.
Some attributes are taken directly from AGS (e.g. raw population and establishment counts), while others are engineered during preprocessing: derived from raw AGS columns through aggregations, ratios, and composite indices (e.g. age-band sums, dependency and sex ratios, density and intensity measures).
Grouped thematically, they cover the structural information shown in \xref{Table}{tab:sources}.
These features are derived from 1,058 raw AGS variables across the seven products.
Additionally, an urbanity flag (binary variable) engineered from population density, nighttime light, and road density was included to help the model differentiate between urban and rural areas.
A total of 572 base features were used for training.

{\tabfont
\renewcommand{\arraystretch}{1.2}
\begin{xltabular}{\linewidth}{@{}Y{0.65} Y{0.95} Y{1.40}@{}}
\caption{\textbf{Applied Geographic Solutions (AGS) data products used to build
PlacePulse embeddings.}}\label{tab:sources}\\
\midrule
Product & Description & What we use from it \\
\midrule
\endfirsthead
\toprule
Product & Description & What we use from it \\
\midrule
\endhead
\toprule
\endfoot
\bottomrule
\endlastfoot
\textbf{Current Year Demographics} & Hundreds of attributes of the population, households, and dwellings, with a special emphasis on the labor force and household income characteristics, including discretionary and disposable income. & Population and age structure, dependency and sex ratios, educational attainment, household composition, income distributions, housing tenure/value/age/type, labor-force participation and unemployment, commuting time and mode, vehicle ownership, race/ethnicity and a diversity index, language spoken at home, marital status, occupation and industry mix, and class of worker.\\
\textbf{Daytime Demographics} & Establishments and employees by detailed industry, and estimates of daytime employment by occupation. & Establishment and employee totals, daytime population, employment and establishment density, retail sales, and per-industry and per-occupation employee and establishment counts. \\
\textbf{Non-Resident Population Estimates} & Estimates of overnight tourist stays by type of travel and type of accommodation, including seasonal estimates. & Visitor and worker flow volumes by type (business, personal, food \& beverage, hotel, etc.) with quarterly distribution ratios and seasonal volatility (quarter-over-quarter change and standard deviation). \\
\textbf{Demographic and Business Dimensions} & Uncorrelated factors summarizing attributes that describe the character of neighborhoods and commercial areas. & AGS's pre-computed dimension indices capturing affluence, age, family status, urbanity, density, sector specialization, and related axes. \\
\textbf{Population Health} & Includes vital statistics, fertility and mortality by age group, life expectancy, chronic health conditions, and health risk indicators.$^{*}$  & Total annual births and deaths, death rates by sex, and life expectancy (total, male, and female).\\
\textbf{Journey to Work} & Summarized commuting flow patterns that includes most employees, expressed as inflow, outflow, and resident workers. & Worker counts capturing both commute ends: residents who live and work in the same cell, residents who commute out, and workers who commute in from elsewhere. \\
\textbf{Natural and Physical Characteristics} & Climate, environmental risks, air quality, and composite measures, as well as terrain and elevation. & Air quality (ozone, PM10, carbon monoxide), annual maximum and minimum temperature, heating and cooling degree days, terrain slope, a walkability index, and a composite risk score. \\ 
\end{xltabular}
\noindent\begin{minipage}{\linewidth}
\notefont Every product is delivered at H3 resolution 7. \\
$^{*}$The chronic-condition and health-risk variables of this product are held out as downstream evaluation targets (\xref{Section}{sec:targets}) and are never model inputs.
\end{minipage}
}

\subsection{Spatial context features: H3 k-ring aggregates}
\label{sec:ring}

The final model is spatially aware: in addition to each cell's own attributes, we included aggregate features computed over its H3 neighborhood on the resolution-7 grid.
For a focal cell $i$, let
\begin{equation}
R_k(i) = \left\{\, j : d_{\mathrm{H3}}(i,j) = k \,\right\}
\label{eq:ring}
\end{equation}
denote the set of cells at H3 grid distance exactly $k$ from $i$, i.e. a ring rather than a cumulative disk, where $d_{\mathrm{H3}}(i,j)$ is the H3 grid distance between cells $i$ and $j$. The focal cell itself is excluded from all neighborhood aggregates to prevent leakage of the center-cell value into the neighborhood summaries, and to ensure that the added features represent the surrounding spatial context rather than the cell's local profile. For each hexagonal H3 cell the first and second rings contain at most $|R_1(i)| = 6$ and $|R_2(i)| = 12$ cells respectively; aggregates are computed over those ring cells present in the grid.

For each of the 572 base features $x$, we added three unweighted neighborhood statistics: the mean over the immediate ring, the sample standard deviation over the immediate ring, and the mean over the second ring.

The ring-$k$ mean is
\begin{equation}
\operatorname{mean}_{k}(x)_i =
\frac{1}{|R_k(i)|}
\sum_{j \in R_k(i)} x_j,
\qquad k \in \{1,2\}.
\label{eq:ringmean}
\end{equation}

The immediate-ring sample standard deviation is
\begin{equation}
\operatorname{sd}_{1}(x)_i =
\sqrt{
\frac{1}{|R_1(i)| - 1}
\sum_{j \in R_1(i)}
\left( x_j - \operatorname{mean}_{1}(x)_i \right)^{2}
}.
\label{eq:ringsd}
\end{equation}

This procedure produces $572 \times 3 = 1{,}716$ spatial context features. Together with the 572 focal-cell attributes, the final model input contains 2,288 features. Each embedding therefore encodes not only a cell's own profile but also the profile of its surrounding H3 neighborhood.

These spatial context features allow the model to distinguish places with identical focal-cell statistics but different neighborhood contexts, such as an affluent enclave bordering a deprived area versus one embedded in a uniformly affluent region. They also encourage embeddings to vary more smoothly across space rather than changing abruptly at cell boundaries, and they provide additional signal for cells whose own attributes are sparse or noisy by borrowing information from nearby cells.

\subsection{Downstream evaluation targets}
\label{sec:targets}

To evaluate the learned embeddings (\xref{Sections}{sec:eval} and~\nref{sec:downstream}) we assemble a battery of target variables from twelve external sources, spanning different spatial units: H3 res-7, ZCTA, census tract and voting precinct (see \xref{Table}{tab:targets}). None of these targets are used in training; they serve only to test how well the embeddings predict held-out quantities. The count targets we construct (POI and cell-tower counts) and the two Startup Cartography Project targets are \texttt{log1p}-transformed as they are extremely right-skewed and span four orders of magnitude.

The evaluation targets were selected to provide both out-of-domain validation and interpretable links to known socio-economic, health, infrastructure, environmental, and political dimensions. Two source relationships are useful for interpreting the results. First, the native-resolution health targets are AGS-modeled disease-prevalence rates, such as diabetes and obesity, whereas the AGS Health inputs used during training are limited to mortality and life-expectancy measures. Performance on these targets therefore tests whether the embeddings capture broader health-related spatial structure beyond the specific health variables seen in training. Second, the socio-economic targets, including income and poverty, are sourced from USDA/Census and are conceptually related to some AGS income attributes in the input set. These targets therefore provide a check on whether the embeddings preserve meaningful socio-economic gradients. We note these relationships in \xref{Table}{tab:targets} and return to them when interpreting the downstream results.

{\tabfont
\renewcommand{\arraystretch}{1.2}
\begin{xltabular}{\linewidth}{@{}Y{1.0} Y{0.45} Y{1.55} Y{0.55} Y{1.0} Y{1.05} Y{1.4}@{}}
\caption{\textbf{Downstream evaluation target sources.}}\label{tab:targets}\\
\midrule
Source & Year & What it measures & Native unit & Evaluation unit & \# targets &
Relation to training inputs$^{*}$ \\
\midrule
\endfirsthead
\toprule
Source & Year & What it measures & Native unit & Evaluation unit & \# targets &
Relation to training inputs$^{*}$ \\
\midrule
\endhead
\toprule
\endfoot
\bottomrule
\endlastfoot
\textbf{AGS Health} & 2025 & Adult disease-prevalence rates: diabetes, obesity, high blood pressure, cancer, high cholesterol, stroke & H3 res-7 & H3 res-7 & 6 & AGS product family and correlated with inputs, but these variables are not model inputs\\
\textbf{CARTO Spatial Features} & 2025 & Human Activity Index (HAI, CARTO derived), nighttime lights (VIIRS), cell-tower density (OpenCellID) & H3 res-8 & H3 res-7, ZCTA (only HAI) & 4 (3 H3 res-7 + 1 ZCTA) & Independent\rev{, except nighttime lights, which enter only indirectly the inputs through the urbanity flag} \\
\textbf{OpenCellID} & 2025 & Cell-tower antennas & Point & ZCTA & 1 & Independent \\
\textbf{Overture Maps} & 2025 & Point-of-interest counts by 11 categories & Point & H3 res-7, ZCTA & 22 (11 H3 res-7 + 11 ZCTA) & Independent; related to AGS establishment counts (by industry rather than place category)\\
\textbf{Google PDFM (\texttt{conus27}) benchmark} & Mixed$^{**}$ & CDC PLACES health indicators; Census/ACS income and home value; nighttime lights, tree cover, elevation, population, density & ZCTA & ZCTA & 28 & Independent; income and health indicators correlate to some AGS inputs, but are not direct model inputs (different source).\\
\textbf{USDA ERS Food Access Atlas} & 2019 & Low-food-access population share, poverty rate, median family income & ZCTA & ZCTA & 3 & Independent; though income and poverty correlate with some AGS inputs (different source) \\
\textbf{NREL / U.S. DOE LEAD} & 2022 & Household energy cost and energy burden & ZCTA & ZCTA & 2 & Independent \\
\textbf{NYT presidential election} & 2024 & Percent Democratic lead & Precinct & Precinct & 1 & Independent \\
\textbf{Opportunity Atlas (Opportunity Insights)} & 2018 & Upward income mobility at age $\sim$35 for children of parents at the 25th income percentile & Tract & Tract & 1 & Independent; linked tax/census records\\
\textbf{FEMA National Risk Index} & 2025 & Social vulnerability to hazard impact, community resilience and capacity to recover & Tract & Tract & 2 & Independent; indicators are built from sociodemographic variables that correlate with inputs\\
\textbf{Social Capital Atlas} & 2022 & Economic connectedness (cross-class bridging), friending clustering & ZIP & ZCTA & 2 & Independent; derived from the Facebook friendship graph \\
\textbf{Startup Cartography Project} & 2016 & Startup formation rate, Regional Entrepreneurship Cohort Potential Index & ZIP & ZCTA & 2 & Independent; derived from business registration records \\
\midrule
\textbf{Total} & --- & --- & --- & H3 res-7 & 20 & --- \\
\textbf{} & --- & --- & --- & ZCTA & 50 & --- \\
\textbf{} & --- & --- & --- & Others & 4 & --- \\
\textbf{} & --- & --- & --- & \textbf{All} & \textbf{74} & --- \\
\end{xltabular}
\vspace{0.5em}
\noindent\begin{minipage}{\linewidth}
\notefont \rev{$^{*}$\textit{Independent} indicates that the target is drawn from a source other than the AGS products used as PlacePulse inputs.} \\
$^{**}$The Google PDFM \texttt{conus27} benchmark combines variables from multiple sources.
We verified the 2022 vintage for the CDC PLACES health indicators, but the available documentation does not specify source years for the remaining variables.
We therefore treat the benchmark as mixed-vintage and partially documented.
\end{minipage}
}

\section{Methods}
\label{sec:methods}

We learn a transferable structural representation for each H3 resolution-7 cell using a self-supervised masked autoencoder trained exclusively on the structural attributes described in \xref{Section}{sec:data}, without downstream task labels. The model compresses the standardized structural feature vector into a compact 256-dimensional embedding and learns to reconstruct masked portions of the input, encouraging the latent representation to capture relationships among the structural characteristics of place. Once trained, the encoder is used as a fixed feature extractor that maps every H3 cell to a single embedding for reuse across diverse downstream tasks. The following subsections describe the model architecture (\xref{Section}{sec:arch}), the masked reconstruction objective (\xref{Section}{sec:loss}), the training procedure (\xref{Section}{sec:training}), and the evaluation protocol used to assess both the intrinsic properties and downstream utility of the learned representations. \xref{Figure}{fig:methods} summarizes the training and evaluation methods.

\begin{figure}[tbp]
\centering
\includegraphics[width=\linewidth]{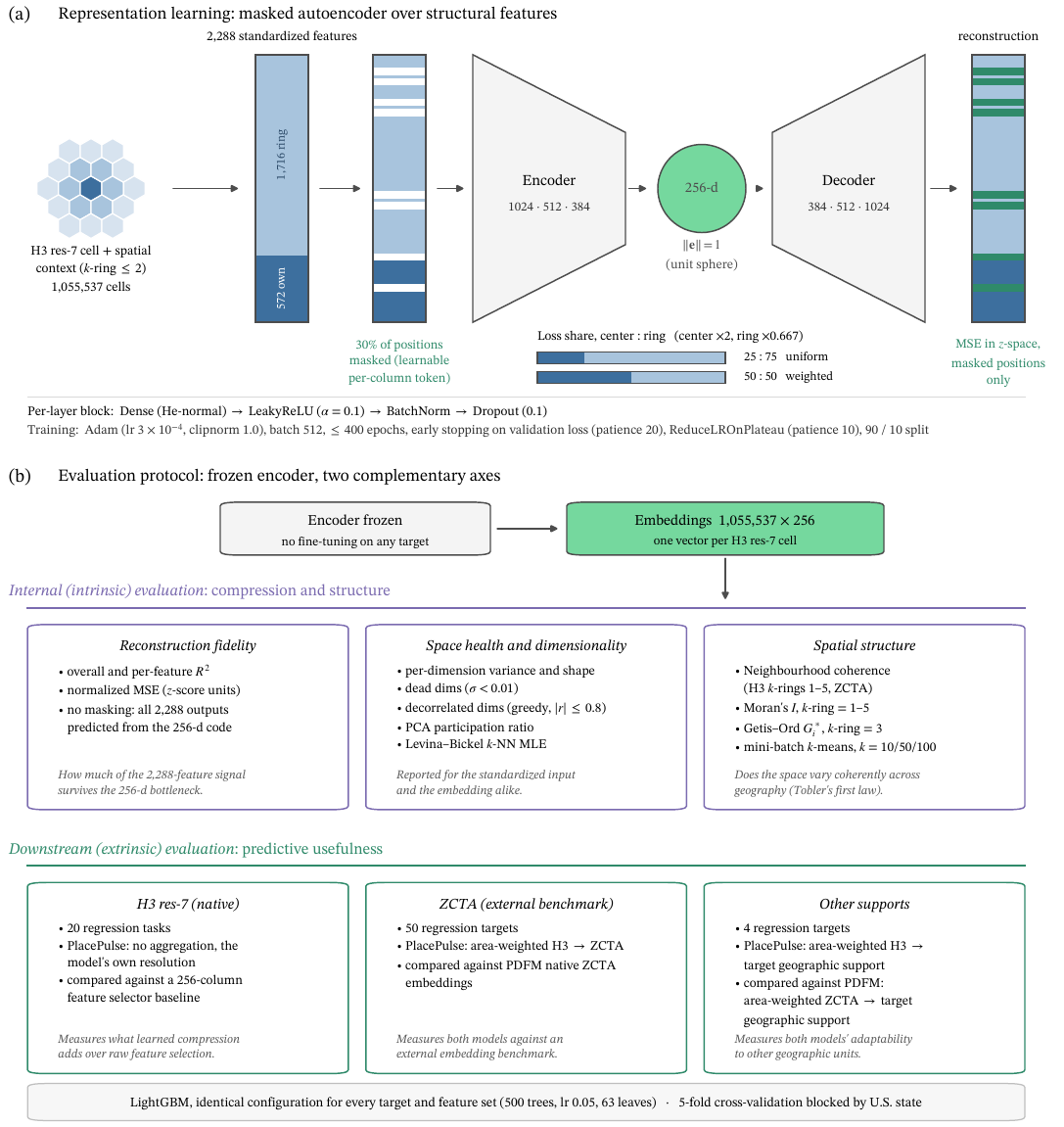}
\caption{\textbf{Overview of the methodology.}
(a) A masked autoencoder compresses 2,288 standardized features per H3 res-7 cell (572 own + 1,716 k-ring aggregates) into a 256-dimensional embedding on the unit sphere, reconstructing only the 30\% of positions masked at each step, with center columns upweighted to offset the 3:1 ring majority.
(b) The frozen encoder is applied to all 1,055,537 cells, and the embeddings are evaluated both intrinsically and as downstream predictors at three spatial supports (see \xref{Section}{sec:eval}).}
\label{fig:methods}
\end{figure}

\subsection{Architecture}
\label{sec:arch}

We use a fully-connected (dense) autoencoder where spatial context is incorporated through the k-ring-aggregate features introduced in \xref{Section}{sec:ring}, allowing the network to learn from both the attributes of each cell and those of its surrounding neighborhood without requiring an explicitly spatial architecture.

The encoder and decoder are symmetric, funnelling the 2,288-dimensional input through progressively narrower layers to the 256-dimensional bottleneck and mirroring that path back out to reconstruct the full vector. Each hidden layer is a standard regularized block (a dense projection followed by LeakyReLU activation, batch normalization, and dropout) and the bottleneck itself is a linear projection followed by an explicit L2-normalization layer, which places every embedding exactly on the unit hypersphere by construction. \xref{Table}{tab:modelconfig} summarizes the configuration.

\begin{table}[tbp]
\centering
\caption{\textbf{Model configuration summary.}}
\label{tab:modelconfig}
\tabfont
\renewcommand{\arraystretch}{1.2}
\begin{tabularx}{\linewidth}{@{}Y{0.55} Y{1.45}@{}}
\toprule
Component & Specification \\
\midrule
Input / output & 2,288 features (572 own + 1,716 ring aggregates) \\
Encoder & Dense [1024, 512, 384] $\rightarrow$ Dense 256 (linear) $\rightarrow$ L2 (unit) normalization \\
Decoder & Dense [384, 512, 1024] $\rightarrow$ Dense 2,288 (linear) \\
Embedding & 256-d, constrained to the unit sphere ($\lVert e \rVert = 1$) \\
Per-layer block & Dense (He-normal init) $\rightarrow$ LeakyReLU ($\alpha = 0.1$) $\rightarrow$ BatchNorm $\rightarrow$ Dropout (0.1) \\
\bottomrule
\end{tabularx}
\end{table}

\subsection{Masked, spatially-weighted reconstruction loss}
\label{sec:loss}

The reconstruction objective incorporates two design choices intended to encourage the learned embeddings to capture relationships among the structural attributes while maintaining a balanced representation of both focal-cell and neighborhood information.

\begin{enumerate}
\item \textbf{\textit{Masked reconstruction.}} During each training step every input position is independently masked with probability 0.30; masked positions are replaced with a learnable per-column ``mask token'' (a trainable vector initialized to zero) before encoding, and the reconstruction loss is computed only on the masked positions. This forces the encoder to infer each masked feature from the others, yielding a representation that captures cross-feature structure rather than copying inputs. Because a fresh random subset is masked every step, the masking also acts as input-level data augmentation, keeping the train--validation gap small. Masking is applied only during training; at inference the encoder receives the complete, unmasked feature vector to generate the final embeddings.

\item \textbf{\textit{Center-vs-ring loss weighting.}} Because ring-aggregate columns outnumber a cell's own columns 3:1 (1,716 vs 572), an unweighted reconstruction loss starves the gradient on the cell's own features and degrades fidelity in dense-urban cells. We therefore apply a two-tier weighting: the cell's own (``center'') columns are weighted $\times 2.0$ and ring columns $\times 0.667$, rebalancing the effective loss contribution from ${\approx}25{:}75$ to ${\approx}50{:}50$ center:ring. This restores the encoder's fidelity to each cell's own attributes while retaining the neighborhood signal that the ring aggregates provide.
\end{enumerate}

The objective is then mean squared error in z-score space, evaluated over masked positions only, with each position scaled by its center/ring group weight. Writing $x_i \in \mathbb{R}^{D}$ for the standardized input of cell $i$ with $D = 2{,}288$, $\hat{x}_i$ for its reconstruction, $m_{id} \in \{0,1\}$ for the mask indicator of position $d$, and $w_d$ for the center/ring group weight of position $d$, the loss over a batch $\mathcal{B}$ is
\begin{equation}
\mathcal{L}
= \frac{\sum_{i \in \mathcal{B}} \sum_{d=1}^{D} m_{id}\, w_d \left( x_{id} - \hat{x}_{id} \right)^{2}}
       {\sum_{i \in \mathcal{B}} \sum_{d=1}^{D} m_{id}\, w_d},
\qquad
w_d =
\begin{cases}
2.0 & \text{if } d \text{ is a center column},\\
0.667 & \text{if } d \text{ is a ring column}.
\end{cases}
\label{eq:loss}
\end{equation}

\subsection{Training}
\label{sec:training}

We train the model to minimize the masked, group-weighted reconstruction objective of \xref{Section}{sec:loss} using the Adam optimizer \citep{kingma2015}, with early stopping on the validation split to select the final weights. The data are split 90\%/10\% into training and validation.

Because the model is used as a fixed feature extractor over a closed, fully-observed grid rather than to predict unseen locations, we do not hold out a separate test set; the validation split governs early stopping and provides a held-out check on reconstruction quality. After training the encoder is applied to all 1,055,537 cells to generate the embeddings. \xref{Table}{tab:trainconfig} lists the full training configuration.

\begin{table}[tbp]
\centering
\caption{\textbf{Model training summary.}}
\label{tab:trainconfig}
\tabfont
\renewcommand{\arraystretch}{1.2}
\begin{tabularx}{\linewidth}{@{}Y{0.55} Y{1.45}@{}}
\toprule
Setting & Value \\
\midrule
Optimizer & Adam, learning rate $3 \times 10^{-4}$, gradient clipping (clipnorm $= 1.0$) \\
Batch size & 512 \\
Max epochs & 400 \\
Early stopping & on validation loss (patience 20, min-$\Delta$ $1 \times 10^{-4}$) \\
LR schedule & ReduceLROnPlateau (patience 10) \\
Train / validation & 90\% / 10\%, random seed 42 \\
\bottomrule
\end{tabularx}
\end{table}

\subsection{Evaluation protocol}
\label{sec:eval}

We evaluate the embeddings along two complementary axes.
Internal (intrinsic) evaluation assesses the representation on its own terms, without reference to any external labels, measuring how faithfully it encodes the input and how well-structured the resulting space is.
Downstream (extrinsic) evaluation measures the embeddings' usefulness as predictors for independent real-world targets.
The two are complementary: a representation may be internally well-conditioned yet of limited predictive value, or vice versa, so we report both.

\textbf{Internal evaluation.} We assess the representation with three families of diagnostics, all computed on the released encoder over the full H3-cell grid.

\begin{enumerate}
\item \textbf{\textit{Reconstruction fidelity.}} Because the model is trained only to reconstruct its input, this directly measures how much of the 2,288-feature signal the 256-d bottleneck preserves (a faithful, near-lossless compression is the precondition for the embedding to be useful downstream).
We report overall and per-feature $R^2$ on the raw feature scale, and normalized MSE in z-score units between the 2,288-d input and its reconstruction.
At evaluation we apply no masking and feed the complete input, so all outputs come from the 256-d code alone.

\item \textbf{\textit{Embedding-space health and dimensionality.}} We evaluate whether the information learned by the encoder is effectively distributed across the embedding space rather than concentrated in a small number of dimensions.
We examine per-dimension variance and distributional shape; the count of \textit{dead} dimensions (per-dimension standard deviation below 0.01); the number of mutually decorrelated dimensions from a greedy test that discards any dimension whose absolute Pearson correlation with an already-retained one exceeds 0.8; and two intrinsic-dimensionality estimators reported for both the standardized input and the embedding: (i) the PCA \textit{participation ratio}, equivalently the effective number of spatial degrees of freedom \citep{bretherton1999} or the $n_2$ effective-dimensionality index \citep{delgiudice2021}, which measures how evenly variance is spread across components; and (ii) the Levina--Bickel maximum-likelihood estimator, a local k-NN measure robust to nonlinear structure \citep{levina2005}.

\item \textbf{\textit{Spatial structure.}} Because the embeddings are meant to describe places, we check that they vary coherently across geography (nearby places should be similar, following Tobler's first law of geography, \citealp{waters2017}) rather than being spatially random, and characterize the scale and form of that structure.
We measure: (i) \textit{neighborhood coherence} for each cell as the cosine similarity between its embedding and the mean embedding of its surrounding neighborhood (an H3 k-ring disk within radius $k$, for $k = 1$--5 for adjacency, or its ZCTA for administrative units), each contrasted with a baseline that compares the cell to a randomly reassigned neighborhood of the same type (significance is assessed with a 99-run permutation test, $p \le 0.01$); and we report both the raw cosine and a mean-centered variant (embeddings minus the global-mean vector), under which the random baseline collapses to ${\sim}0$ so the statistic isolates excess local similarity beyond the national average; (ii) \textit{global spatial autocorrelation} by computing Moran's $I$ \citep{moran1950} over H3 k-ring adjacency with row-standardized weights, computed on the leading PCA components and on each of the 256 dimensions, with permutation-based significance and a correlogram over ring distances $k = 1$--5; (iii) \textit{local hotspot structure} through a Getis--Ord $G_i^{*}$ analysis \citep{getis1992, ord1995} with uniform weights over H3 k-ring-3 neighborhoods, combining the 256 per-dimension $G_i^{*}$ z-scores into a single per-cell statistic through a standardized sum \citep{stouffer1949, whitlock2005}; and (iv) \textit{mini-batch K-means cluster structure} \citep{sculley2010} at $k = 10/50/100$, measuring size balance, cluster-size entropy $H$ \citep{shannon1948} and its effective-cluster equivalent $2^{H}$, the number of equally-sized clusters carrying the same entropy \citep{hill1973}, and within-cluster sum of squares.
\end{enumerate}

\textbf{Downstream evaluation.} The embeddings are evaluated as predictors across a battery of supervised regression tasks (20 at native H3 res-7, 50 at ZCTA, and 4 at other geographic support levels) in every case using 5-fold cross-validation with gradient-boosted decision trees (LightGBM; \citealp{ke2017}). All targets use an identical LightGBM configuration (\texttt{n\_estimators} $= 500$, \texttt{learning\_rate} $= 0.05$, \texttt{num\_leaves} $= 63$); no per-target or per-feature-set tuning, so differences reflect the representation, not the estimator.

We report the evaluation in three settings:

\begin{enumerate}
\item \textbf{\textit{At the native resolution of the embeddings (H3 res-7).}} Targets available at H3 res-7 are predicted directly from the 256-d cell embeddings, quantifying how well the structural representation predicts each target at the model's native, fine resolution.
Because no external open-source embedding product is available at H3 resolution 7, we compare PlacePulse to an internal dimensionality-matched, target-blind feature-selection \textit{baseline}.
This baseline is constructed from the original input features: the 2,288 standardized inputs are grouped using agglomerative clustering and cut into 256 clusters, and one medoid feature is retained from each cluster. The medoid is defined as the feature with the smallest mean distance to the other features in its cluster. The resulting 256-feature set is selected once, without access to any target labels, and is used unchanged across all prediction tasks, preventing label leakage into any cross-validation fold. The feature-selection baseline is deliberately strong: it retains interpretable AGS columns, reduces redundancy among correlated inputs, and broadly covers the original input space.
The comparison therefore asks whether the encoder earns its 256 learned dimensions over a simple target-blind selection of 256 representative input features.

\item \textbf{\textit{At ZCTA resolution, against external baselines.}} Targets defined at ZCTA are predicted after aggregating PlacePulse embeddings from H3 res-7 up to ZCTA through an area-weighted mean, each cell contributing in proportion to its area of overlap with the ZCTA. This lets us benchmark against external baselines, notably Google's Population Dynamics Foundation Model \citep[PDFM;][]{agarwal2024}, which is published natively at ZCTA/county and therefore enters the comparison with no aggregation of its own. Both models are thus evaluated on the same ZCTA prediction task, but only PlacePulse is aggregated to get there. \rev{The comparison is therefore conservative for our model: PDFM is evaluated at the native resolution it was trained on, whereas PlacePulse is evaluated through an area-weighted mean of embeddings learned for H3 cells.} \rev{Note that the 5-fold cross-validation and LightGBM configuration differ from the original PDFM evaluation \citep{agarwal2024}, which used different LightGBM hyperparameters and a single holdout of 20\% of states, so absolute $R^2$ values are not directly comparable with those reported there.}

\item \textbf{\textit{At other target resolutions.}} The same target-resolution procedure extends to spatial units other than ZCTA.
Here neither model is native to the target unit, so both PlacePulse and PDFM are area-weighted to the target resolution (i.e. the voting precinct).
\rev{The transfer is not symmetric, because the two native supports differ in size and in how they align with the target units.}
\rev{Because ZCTAs, tracts and precincts are all built from census blocks, most target units fall almost entirely inside a single ZCTA.}
\rev{PDFM therefore transfers cleanly but at a coarse resolution (all units inside one ZCTA share one embedding), whereas PlacePulse transfers at a finer resolution but mixes values across neighbouring units.}
\rev{Because H3 is a regular hexagonal grid, its cell boundaries are independent of census geography and rarely coincide with tract or precinct boundaries: units that straddle several cells receive a blend of their neighbours' embeddings, while units small enough to fall entirely inside one cell receive that cell's embedding unchanged, shared with every other unit in the same cell.}
\end{enumerate}

Because all reported targets have nationwide coverage, we use \textit{spatially blocked cross-validation} throughout, grouping observations by U.S. state so that all cells from a given state are assigned to the same fold. Holding out whole states reduces leakage from spatial autocorrelation between nearby training and test observations and provides a more conservative estimate of geographic generalization. Folds are constructed to contain approximately similar numbers of H3 cells. We report the average and standard deviation $R^2$ across the 5 held-out folds for each downstream task.

\section{Results}
\label{sec:results}

We organize results along the two axes presented in \xref{Section}{sec:eval}. \xref{Section}{sec:internal} reports the internal evaluation (reconstruction fidelity, embedding geometry, and spatial/cluster structure) over the full 1,055,537-cell grid. \xref{Section}{sec:downstream} reports the downstream evaluation against PDFM and a feature-selector baseline model.

\subsection{Internal evaluation}
\label{sec:internal}

The intrinsic evaluation examines the properties of the learned embedding space. We first assess reconstruction fidelity to quantify how accurately the compressed representation preserves the original structural information. We then analyze the statistical and spatial properties of the embedding space through complementary measures of representation geometry and spatial organization.

\subsubsection{Reconstruction fidelity}
\label{sec:recon}

The autoencoder reconstructs the standardized input with average $R^2 = 0.853$, corresponding to a normalized MSE of 0.147 in z-score units, where 1.0 is a per-feature mean predictor and 0.0 is perfect reconstruction. The per-feature $R^2$ distribution (\xref[a]{Figure}{fig:quality}) is tight and uniformly positive: median $R^2 = 0.887$, with $p_{10} = 0.686$ and $p_{90} = 0.968$. None of the 2,288 features has $R^2 < 0$, so no feature is reconstructed worse than its own mean. Residuals, computed as observed minus reconstructed values, are centered close to zero, indicating no systematic aggregate reconstruction bias; the remaining error is concentrated in feature-specific tails rather than broad miscalibration.

\begin{figure}[tbp]
\centering
\includegraphics[width=\linewidth]{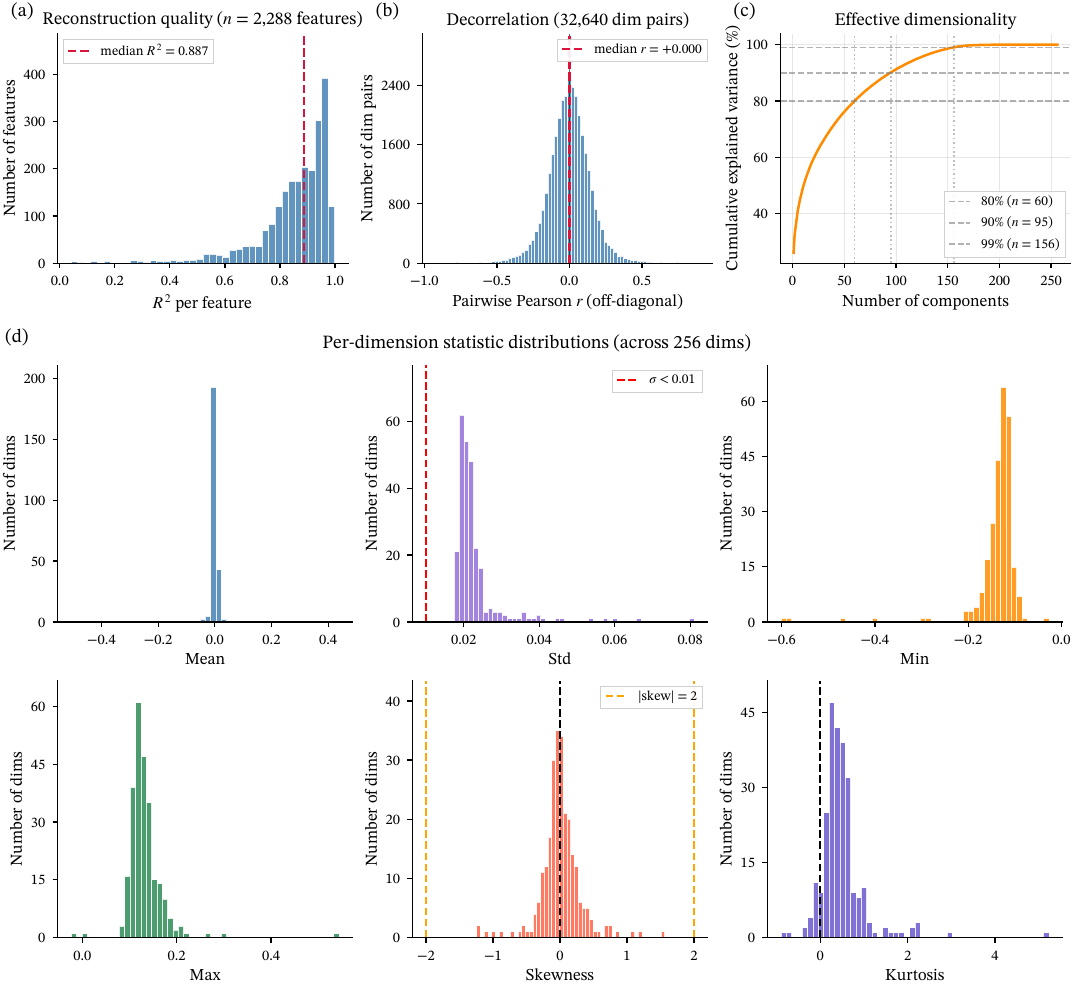}
\caption{\textbf{Internal quality diagnostics of the 256-dimensional embeddings, computed over all 1,055,537 H3 res-7 cells.}
(a) Distribution of per-feature reconstruction $R^2$ across the 2,288 input features; dashed line, median.
(b) Distribution of Pearson correlation coefficients across all 32,640 off-diagonal embedding-dimension pairs; dashed line, median.
(c) Cumulative PCA explained variance against number of components; dashed lines mark the 80\%, 90\% and 99\% levels.
(d) Distributions across the 256 embedding dimensions of six per-dimension summary statistics, each computed over all cells: mean, standard deviation, minimum, maximum, skewness, and excess kurtosis.
Red dashed line, the 0.01 dead-dimension threshold (\xref{Section}{sec:eval}); orange lines, $|\text{skewness}| = 2$.}
\label{fig:quality}
\end{figure}

\subsubsection{Embedding space health and dimensionality}
\label{sec:health}

Beyond preserving the structural information contained in the input features, a useful embedding should make effective use of its representational capacity. We therefore examined whether the learned 256-dimensional space is well-conditioned and free of the degeneracies that commonly afflict autoencoder bottlenecks.

\textbf{\textit{Unit norm by construction.}} Every embedding lies exactly on the unit hypersphere: the L2-normalization layer (\xref{Section}{sec:arch}) enforces $\lVert e \rVert = 1$ by construction, and the measured norms confirm this exactly (mean 1.0000, standard deviation 0.0000 across all 1,055,537 cells).
A direct and useful consequence is that cosine similarity between any two embeddings reduces to their dot product, which makes similarity and nearest-neighbor operations inexpensive at national scale.

\textbf{\textit{High effective rank.}} The distribution of correlation coefficients across the off-diagonal embedding-dimension pairs is symmetric and tightly concentrated at zero (median $r = 0.000$, with 99\% of pairs below $|r| = 0.47$), and the greedy decorrelation test retains 251 of the 256 dimensions, so only five are strongly linearly correlated with another: the bottleneck spans a broad effective subspace rather than collapsing onto a few axes (see \xref[b]{Figure}{fig:quality}), consistent with the intrinsic-dimensionality analysis below.
This is a linear redundancy diagnostic; because both the encoder and the downstream models (\xref{Section}{sec:downstream}) are nonlinear, even the few linearly-correlated dimensions may still carry complementary information that nonlinear models can exploit.

\textbf{\textit{Intrinsic dimensionality.}} We report the PCA participation ratio and the Levina--Bickel MLE estimator (\xref{Section}{sec:eval}) for both the 2,288-d standardized input and the 256-d embedding, computed on the full grid with the model's own train-split standardization so the two are directly comparable (\xref{Table}{tab:variance}).

\begin{table}[tbp]
\centering
\caption{\textbf{Variance structure of the input vs. the embedding.}}
\label{tab:variance}
\tabfont
\renewcommand{\arraystretch}{1.2}
\begin{tabularx}{\linewidth}{@{}Z r r@{}}
\toprule
Measure & Input (2,288-d) & Embedding (256-d) \\
\midrule
PCA participation ratio (evenness of variance spread) & 6.6 & 12.6 \\
Levina--Bickel MLE (nonlinear degrees of freedom) & 32.3 & 25.2 \\
Variance in top-3 PCA components (coverage) & 47.8\% & 35.7\% \\
\bottomrule
\end{tabularx}
\end{table}

First, the autoencoder decorrelates a highly redundant input.
The 2,288 ring-augmented features are dominated by a few directions (top-3 PCA components alone carry 47.8\% of variance, since each ring aggregate correlates strongly with its center feature), giving a low participation ratio of 6.6.
The embedding nearly doubles the participation ratio (to 12.6) while its top-3 coverage drops to 35.7\%.
These are two views of the same effect: variance is spread more evenly across directions rather than piled into a few.
Second, the embedding retains most of the input's nonlinear degrees of freedom (MLE 25.2 vs 32.3, ${\approx}78\%$) while both estimates sit well below the 256-d bottleneck, so the representation is compact rather than capacity-bound.
The full cumulative explained variance distribution of the embedding is illustrated in \xref[c]{Figure}{fig:quality}. Rather than collapsing onto a few dominant axes, the embedding distributes its variance across many components (60 principal components are needed to reach 80\% of the total variance, 95 for 90\%, and 156 for 99\%).

\textbf{\textit{No dead or degenerate dimensions.}} No dimension is collapsed or unused and the model uses its full capacity: as per \xref[d]{Figure}{fig:quality}, the per-dimension standard deviations are small but tightly clustered (mean 0.023, range 0.018--0.081; above the 0.01 dead-dimension threshold).
The small scale is a structural effect of unit-normalization; the key point is that all 256 dimensions are nonzero and comparable in magnitude, so the bottleneck is used broadly rather than collapsing onto a few active dimensions.
Additionally, the embedding dimension means are tightly concentrated at zero; the small, bounded min/max ranges are consistent with the unit-norm (L2) embedding geometry; and all dimensions are approximately symmetric ($|\text{skewness}| < 2$, orange lines) with only mild, mostly positive excess kurtosis, indicating well-behaved, near-Gaussian marginals rather than heavy-tailed or degenerate ones.

\subsubsection{Spatial structure}
\label{sec:spatial}

To assess whether the embeddings preserve local geographic structure, we compare embedding similarity across adjacent, co-located, and randomly selected cell pairs. We find that the embeddings are geographically smooth: nearby and co-located cells exhibit substantially higher similarity than random pairs. These results suggest that the learned representation captures the spatial autocorrelation inherent in the input attributes, where nearby locations tend to have similar demographic and business composition, consistent with Tobler's first law of geography \citep{waters2017}. The use of k-ring-aggregate inputs further strengthens this spatial structure by incorporating information from each cell's surrounding context.

\textbf{\textit{Neighborhood coherence.}} We evaluate whether the embeddings preserve local spatial structure by comparing each cell's embedding with the mean embedding of its surrounding neighborhood, and contrasting this with a spatially permuted baseline (\xref{Figure}{fig:coherence}).

\begin{figure}[tbp]
\centering
\includegraphics[width=\linewidth]{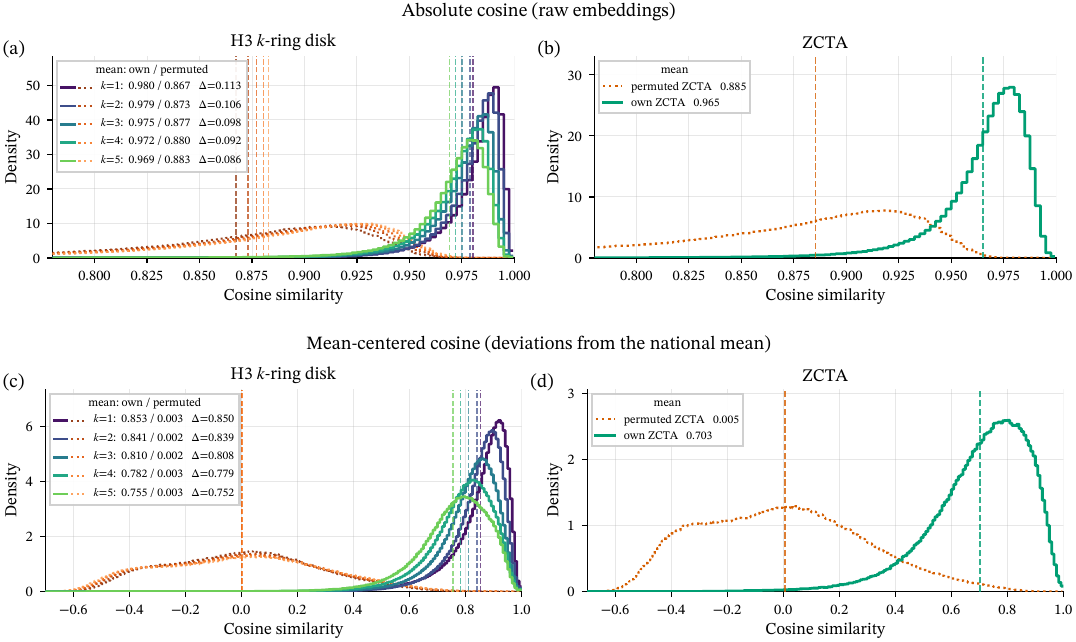}
\caption{\textbf{Neighborhood coherence of the embeddings as the density of the per-cell cosine similarity between a cell's embedding and the mean embedding of its neighborhood.}
Rows are the two framings: (a, b) the PlacePulse embeddings, (c, d) the same embeddings minus the global mean vector.
Columns are the two neighborhood definitions: (a, c) the cumulative H3 k-ring disk for $k = 1$--5, focal cell excluded, over all 1,055,537 cells; (b, d) the cell's own ZCTA area-weighted mean, focal cell included, over the 1,001,075 cells assigned to one of 30,283 ZCTAs.
Solid step curves are own-neighborhood similarities.
Dotted curves are the permuted baseline, in which the same cell is compared with a randomly reassigned neighborhood of the same type.
In (a, c) color encodes the ring radius, purple to green for $k = 1$ to 5, and all five permuted baselines are orange.
Dashed vertical lines mark the means.}
\label{fig:coherence}
\end{figure}

For H3 adjacency (the $k = 1$ ring), the mean cosine similarity between a cell and its neighborhood mean is 0.980 ($p_{10} = 0.962$, $p_{90} = 0.994$), versus 0.867 for a permuted-neighborhood baseline (the same cell compared to a random cell's neighborhood), a $1.13\times$ lift; coherence decays only gradually with radius, to 0.969 at $k = 5$ (\xref[a]{Figure}{fig:coherence}).
For ZCTAs, the mean cell-to-ZCTA-mean similarity is 0.965 ($p_{10} = 0.939$, $p_{90} = 0.986$), versus 0.885 for a permuted-ZCTA baseline, a $1.09\times$ lift (\xref[b]{Figure}{fig:coherence}).
All contrasts are overwhelmingly significant under a 99-permutation null ($p \le 0.01$).
The high absolute similarities indicate that the embedding space is geographically smooth, supporting the use of area-weighted mean aggregation for downstream enrichment (\xref{Section}{sec:downstream}).
However, raw cosine similarities are partly compressed by a shared ``national-average'' direction in the L2-normalized embeddings.
After mean-centering, this common-direction floor is removed: the permuted baselines collapse to approximately zero, while own-neighborhood similarity remains high at 0.853 for immediate H3 neighbors and 0.703 for ZCTAs (\xref[c]{Figure}{fig:coherence} and \xref[d]{Figure}{fig:coherence}).
This confirms that the embeddings capture local spatial coherence beyond the global average structure.
ZCTA aggregation is therefore justified because it averages coherent local vectors rather than unrelated embeddings, although it is not lossless.
Quantifying this aggregation loss is left for future work.

\textbf{\textit{Spatial autocorrelation (Moran's $I$).}} In the 256-dimensional embedding space, spatial autocorrelation is strong and highly significant.
For adjacent units $k = 1$, all but one of the 256 embedding dimensions have Moran's $I > 0.50$ (the weakest is 0.495 and the mean is 0.70) with permutation $p < 0.005$ against the null expectation $E[I] \approx 0$.
Thus, spatial structure is present across the embedding coordinates rather than being confined to a small subset of dimensions.
We next examined the same structure after rotating the embedding with PCA.
The dominant principal component, which explains 26\% of the variance, is especially spatially autocorrelated, with Moran's $I = 0.93$ at $k = 1$, showing that spatial organization is distributed throughout the representation.
Autocorrelation is not uniform across all PCA directions: some lower-variance components show weaker spatial autocorrelation, consistent with these components capturing finer-grained, more localized, or more specialized variation after broad-scale gradients have been accounted for.
The correlogram over ring distances $k = 1$--5 shows gradual monotonic decay rather than a sharp drop (\xref[a--b]{Figure}{fig:autocorr}).
For PC1, Moran's $I$ declines from 0.93 to 0.83, while the mean across embedding dimensions declines from 0.70 to 0.50.
Autocorrelation therefore remains high beyond the two-ring radius of the model's own inputs, suggesting that the learned representation reflects broad-scale spatial organization rather than merely the local ring-aggregate features fed to the encoder.

\textbf{\textit{Local hotspot structure (Getis--Ord $G_i^{*}$).}} As a local companion to the global Moran's $I$, the Getis--Ord $G_i^{*}$ hotspot analysis confirms pronounced local clustering: the combined statistic ranges from $-18.3$ to $+24.5$, and 55.6\% of the cells fall in statistically significant clusters (two-sided $p < 0.05$).
This local view (see \xref[c]{Figure}{fig:autocorr}) reinforces the global autocorrelation result: the embedding space organizes into coherent regional structures rather than being distributed at random.
Note that because embedding dimensions have arbitrary orientation, the sign of the aggregate is not semantically directional; the statistic is used here as a spatial-clustering diagnostic, not a labeled hot/cold map.

\begin{figure}[tbp]
\centering
\includegraphics[width=\linewidth]{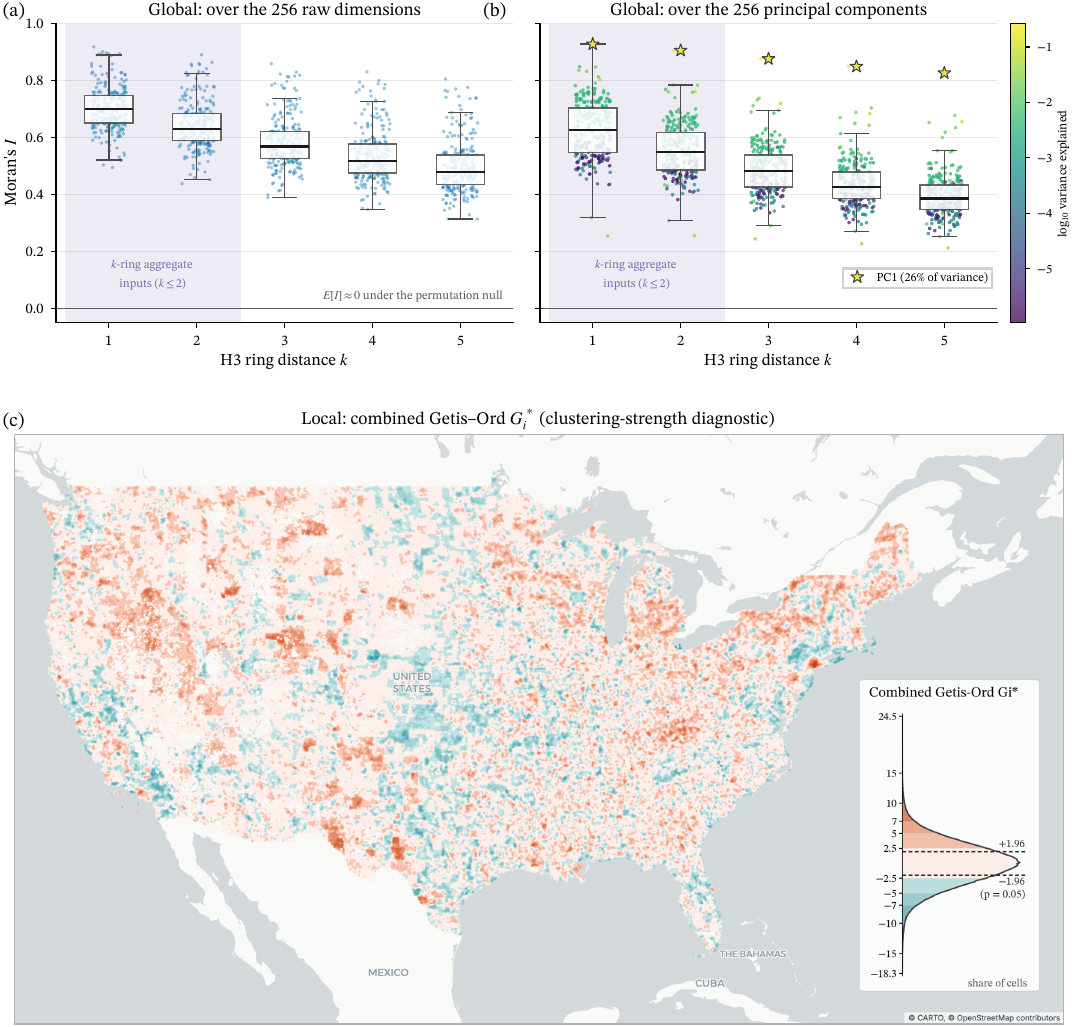}
\caption{\textbf{Spatial autocorrelation of the embeddings across all 1,055,537 H3 resolution-7 cells.}
(a) Moran's $I$ for the 256 original embedding dimensions, using row-standardized H3 ring-neighborhood weights for $k = 1$--5.
Boxes show the median and interquartile range across dimensions, with individual dimensions shown as jittered points.
(b) The same analysis for the 256 principal components; points are colored by $\log_{10}$ variance explained, and the star marks PC1.
In both panels, the shaded band marks $k \leq 2$, the radius of the encoder's ring-aggregate inputs, and the horizontal line marks $E[I] \approx 0$.
(c) Combined Getis--Ord $G_i^{*}$, as defined in \xref{Section}{sec:eval}, over uniform-weight H3 ($k = 3$) ring neighborhoods.
Values are binned into 11 classes, with teal negative and red positive.
The inset gives class edges, cell shares, and dashed $\pm 1.96$ reference lines for a two-sided $p = 0.05$ threshold.
Because embedding orientation is arbitrary, the sign is not directional.}
\label{fig:autocorr}
\end{figure}

\textbf{\textit{Cluster structure.}} As seen in \xref{Figure}{fig:clusters}, K-means partitions of the embedding space are interpretable rather than arbitrary.
At $k = 10$ the partition orders locations by settlement density, from urban core suburbs to sparse rural land, while also capturing patterns beyond differences in density alone (see \xref[a]{Figure}{fig:clusters}).
\xref[b]{Figure}{fig:clusters} illustrates how finer partitions mostly refine this structure rather than reorganizing, which indicates substantial multiscale consistency despite K-means not being a hierarchical algorithm.
In particular, a typical $k = 100$ cluster draws a median of 70.8\% of its cells from a single $k = 10$ parent cluster, and a typical $k = 50$ cluster draws a median of 68.4\% from its $k = 10$ parent, both well above the 18.5\% chance baseline.
\xref{Table}{tab:clusters} shows how within-cluster inertia declines gradually with increasing $k$, from 102,239 to 89,946 to 84,510 for $k = 10 \rightarrow 50 \rightarrow 100$, only a 17\% reduction despite a tenfold increase in cluster count.
This pattern is consistent with high neighbor coherence (\xref{Section}{sec:spatial}) and modest intrinsic dimensionality (\xref{Section}{sec:health}).
The partitions also avoid a dominant mode: the largest cluster contains 18.5\% of cells at $k = 10$ and only 2.2\% at $k = 100$, while cluster-size entropy reaches 94--98\% of the value a perfectly even split would give.
Fine-grained niche groups still emerge at higher $k$, with the smallest $k = 100$ cluster containing approximately 700 cells.
Refer to \xref{Appendix}{app:clustering} for more details.

\begin{table}[tbp]
\centering
\caption{\textbf{Mini-batch K-means cluster structure over the 256-dimensional embeddings.}
$H$ is the Shannon entropy of the cluster-size shares, its maximum $\log_2 k$ attained at equal sizes; $2^{H}$ is the equivalent number of equally-sized clusters; inertia is the within-cluster sum of squares.}
\label{tab:clusters}
\tabfont
\renewcommand{\arraystretch}{1.2}
\begin{tabular*}{\linewidth}{@{\extracolsep{\fill}}r l l l r r r@{}}
\toprule
$k$ & Largest cluster & Smallest cluster & Entropy $H$ & $H$ / max & Effective clusters & Inertia \\
\midrule
10 & 195,375 (18.5\%) & 36,385 (3.45\%) & 3.14 (max 3.32) & 94.5\% & 8.8 & 102,239 \\
50 & 39,340 (3.7\%) & 4,939 (0.47\%) & 5.50 (max 5.64) & 97.5\% & 45.4 & 89,946 \\
100 & 22,963 (2.2\%) & 696 (0.07\%) & 6.48 (max 6.64) & 97.5\% & 89.1 & 84,510 \\
\bottomrule
\end{tabular*}
\end{table}

\begin{figure}[tbp]
\centering
\includegraphics[width=\linewidth]{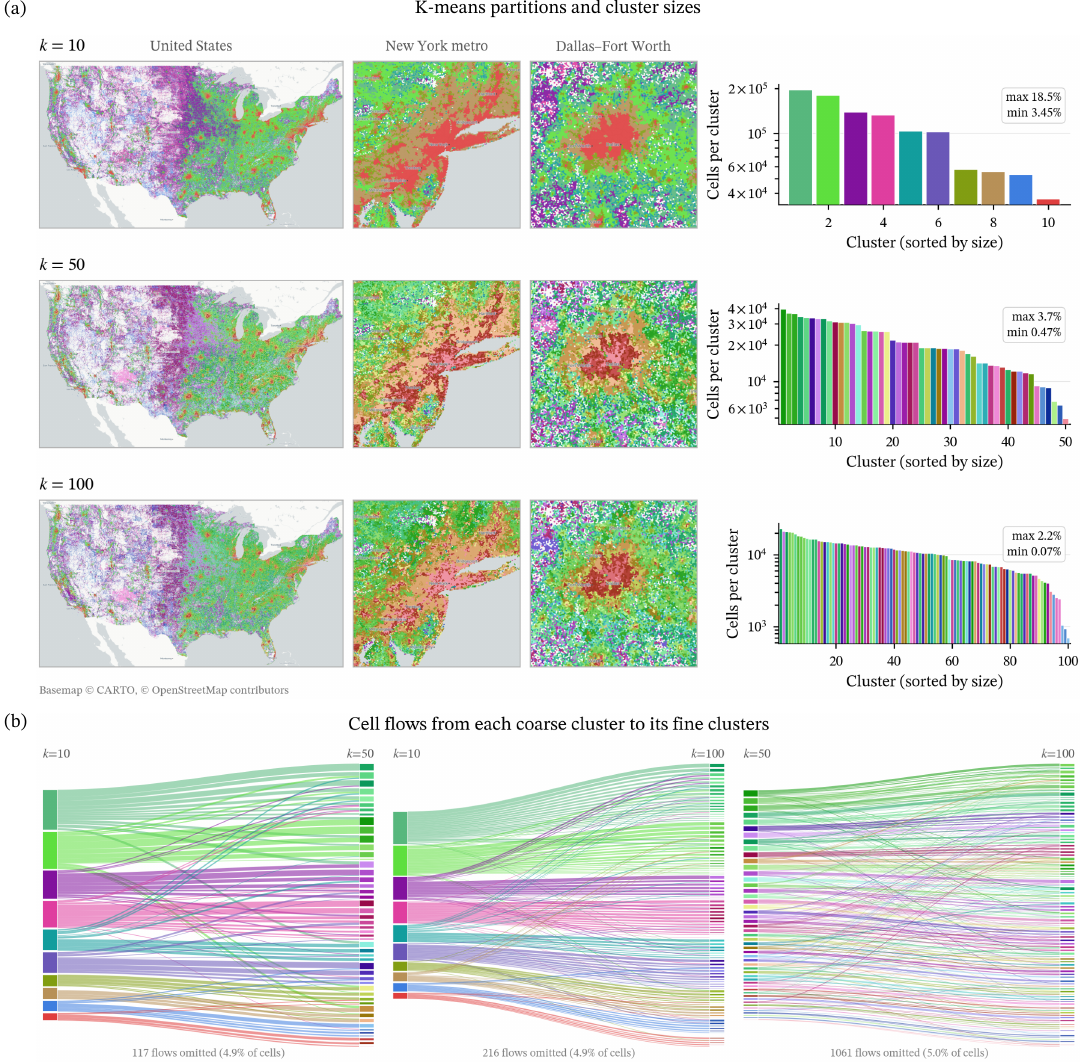}
\caption{\textbf{K-means cluster structure over all H3 res-7 cells.}
Color identifies the cluster and is shared across all plots: clusters close in embedding space take close colors, and at $k = 50$ and 100 each cluster takes the hue of the $k = 10$ cluster holding most of its cells and varies only in lightness.
(a) Partitions at $k = 10$ (top), 50 (middle) and 100 (bottom).
Left to right: the conterminous United States, a New York metro close-up, a Dallas--Fort Worth close-up, and the cluster sizes sorted largest to smallest on a log axis, annotated with the largest and smallest share.
(b) Cell flows from the coarse to the fine partition, for $k = 10 \rightarrow 50$, $10 \rightarrow 100$ and $50 \rightarrow 100$ (left to right).
Ribbon width is proportional to the number of cells.
Only the largest ribbons carrying 95\% of cells are drawn, and each diagram states how many flows it omits.}
\label{fig:clusters}
\end{figure}

\subsection{Downstream predictive utility}
\label{sec:downstream}

The downstream evaluation examines the utility of the learned embeddings in applied predictive tasks.
We assess whether the representations capture information that is useful for modeling real-world outcomes across different domains and spatial resolutions.
This analysis provides an external measure of how well the embeddings generalize beyond their training objectives and support practical geospatial modeling workflows.

\subsubsection{Native resolution (H3 res-7)}
\label{sec:down-h3}

To our knowledge, no comparable public embedding product is natively available at H3 resolution 7, so the native-resolution comparison is limited to PlacePulse and the baseline representation defined in \xref{Section}{sec:eval}.
\xref{Table}{tab:down-h3} summarizes performance across the 20 native-resolution targets listed in \xref{Table}{tab:targets}.
\xref{Figure}{fig:down-h3} and \xref{Table}{tab:b1} give the full per-target results.

\begin{table}[tbp]
\centering
\caption{\textbf{Downstream performance at native H3 res-7 resolution (PlacePulse vs. the target-blind, dimensionality-matched baseline).}
The average per-fold $R^2$ of the tasks in each category is reported; see \xref{Table}{tab:b1} for the uncertainty of each individual task.}
\label{tab:down-h3}
\tabfont
\renewcommand{\arraystretch}{1.2}
\begin{tabularx}{\linewidth}{@{}Y{0.73} Y{1.27} r r r@{}}
\toprule
Category & What it covers & \# & PlacePulse $R^2$ & Baseline $R^2$ \\
\midrule
\textbf{Public Health} & Adult health-risk prevalence: diabetes, obesity, high blood pressure, cancer, high cholesterol, stroke. & 6 & 0.707 & 0.738 \\
\textbf{Points of Interest} \textit{(log space)} & POI counts by type: total plus retail, food \& drink, healthcare, offices, finance, leisure, education, public services, tourism, transport. & 11 & 0.772 & 0.724 \\
\textbf{Population Activity} & Human-footprint proxies: human activity index, nighttime lights (raw VIIRS radiance), cell-tower density. & 3 & 0.801 & 0.788 \\
\bottomrule
\end{tabularx}
\end{table}

Public Health is the category most closely aligned with the training distribution.
As noted in \xref{Section}{sec:targets}, these targets are AGS-modeled disease-prevalence rates from the same product family as the mortality and life-expectancy variables used during training, although they are not themselves training inputs.
PlacePulse trails the baseline modestly in this category, with an average $R^2$ of 0.707 versus 0.738.
This suggests that the baseline retains an advantage on health-adjacent variables, whereas PlacePulse allocates its fixed-dimensional capacity across a broader set of place attributes.

PlacePulse performs better in the two categories with no direct relationship to the training inputs.
For Points of Interest, PlacePulse outperforms the baseline by a substantial margin: $R^2$ is 0.772 versus 0.724.
For Population Activity, PlacePulse also leads, though by a smaller margin, with $R^2$ of 0.801 versus 0.788.
The strongest individual result is for the Human Activity Index, where PlacePulse achieves $R^2 = 0.930$.
These results suggest that PlacePulse learns a more general representation of place structure, retaining functional and activity-related information beyond what is captured by a compact target-blind subset of the original inputs.

\begin{figure}[tbp]
\centering
\includegraphics[width=0.675\linewidth]{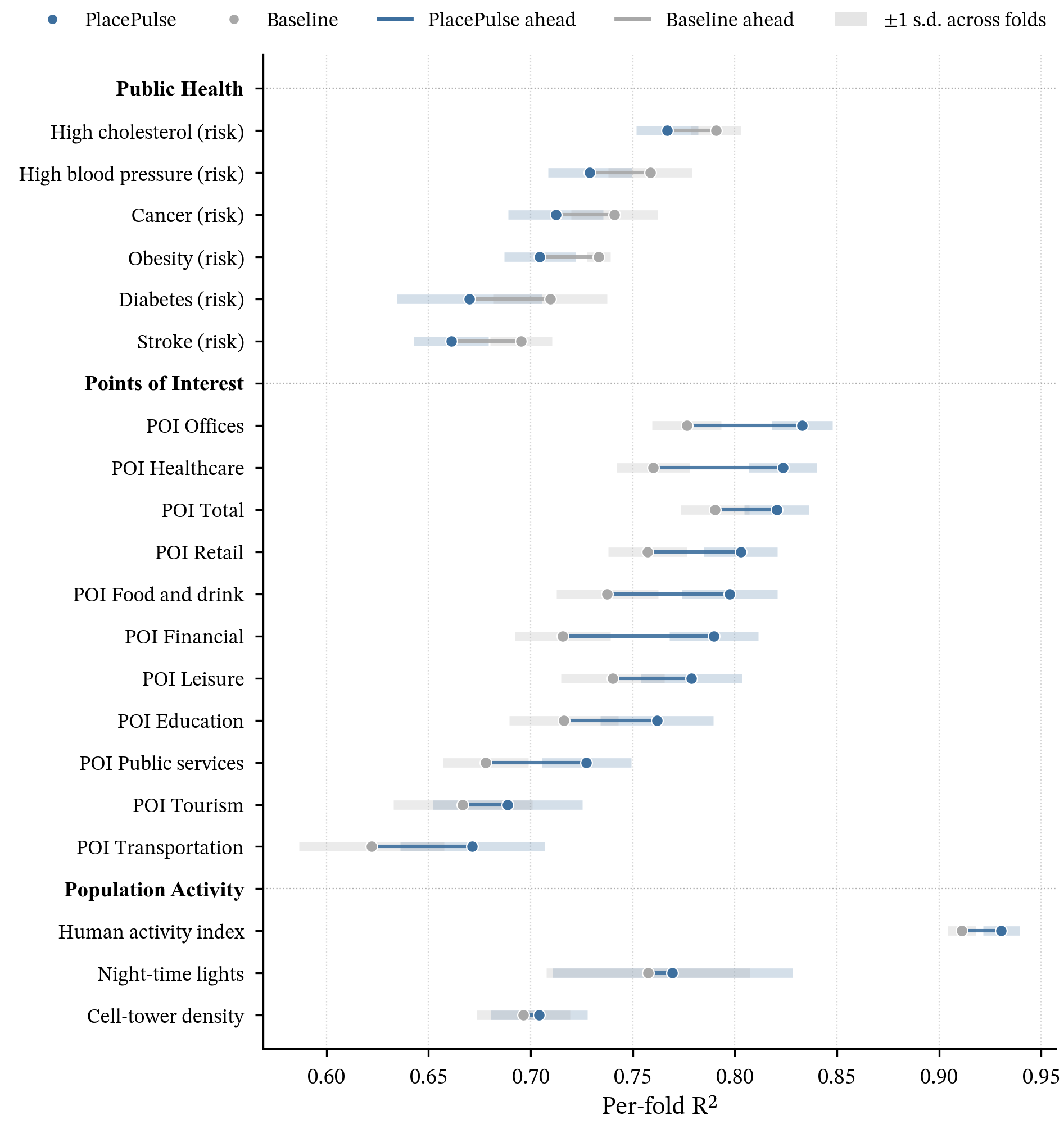}
\caption{\textbf{Downstream predictive performance at native H3 res-7 resolution.}
PlacePulse (256-d) vs. the target-blind, dimensionality-matched baseline, per target within each of the three native-resolution categories, under 5-fold spatially-blocked cross-validation.
Each row is one target: the two dots are that target's per-fold mean $R^2$ under the two representations, the connector takes the color of the leading one, and the shaded band spans $\pm1$ standard deviation across the 5 held-out folds.}
\label{fig:down-h3}
\end{figure}

\subsubsection{External baseline's resolution (ZCTA)}
\label{sec:down-zcta}

At ZCTA resolution, PlacePulse embeddings are evaluated after aggregation from H3 resolution 7 to ZCTAs using an area-weighted mean, as described in \xref{Section}{sec:eval}.
They are compared with Google's Population Dynamics Foundation Model \citep[PDFM;][]{agarwal2024}, which is published natively at ZCTA resolution.
\rev{This comparison is conservative for PlacePulse, because PDFM is evaluated at the native resolution it was trained on, whereas PlacePulse is evaluated at ZCTA through an area-weighted mean of embeddings learned for H3 cells.}
Targets span seven categories, summarized in \xref{Table}{tab:down-zcta} and fully described in \xref{Figure}{fig:down-zcta} and \xref{Table}{tab:b2}.

\begin{table}[tbp]
\centering
\caption{\textbf{Downstream performance at ZCTA resolution (PlacePulse vs. PDFM).}
The average per-fold $R^2$ of the tasks in each category is reported; see \xref{Table}{tab:b2} for the uncertainty of each individual task.}
\label{tab:down-zcta}
\tabfont
\renewcommand{\arraystretch}{1.2}
\begin{tabularx}{\linewidth}{@{}Y{0.73} Y{1.27} r r r@{}}
\toprule
Category & What it covers & \# & PlacePulse $R^2$ & PDFM $R^2$ \\
\midrule
\textbf{Public Health} & Health indicators: chronic-disease prevalence, health behaviors, self-rated physical and mental health, preventive-care access. & 21 & 0.610 & 0.573 \\
\textbf{Economic status} & Economic standing and hardship: median family and household income, median home value, poverty rate, energy burden and cost, low-food-access population share. & 7 & 0.633 & 0.605 \\
\textbf{Population Activity} \textit{(mixed scales)} & Population, population density, nighttime lights; human activity index and cell-tower density. & 5 & 0.863 & 0.855 \\
\textbf{Physical Geography} & Natural land signals: mean elevation and tree-canopy share. & 2 & 0.628 & 0.744 \\
\textbf{Points of Interest} \textit{(log space)} & POI counts by type, the same 11 classes as at H3. & 11 & 0.784 & 0.906 \\
\textbf{Social capital} & Cross-status economic connectedness and friendship clustering. & 2 & 0.715 & 0.713 \\
\textbf{Entrepreneurship} \textit{(log space)} & Startup formation rate and regional entrepreneurship cohort potential. & 2 & 0.765 & 0.822 \\
\bottomrule
\end{tabularx}
\end{table}

\rev{On average, }PlacePulse \rev{achieves a higher $R^2$ than} PDFM in the Public Health (0.610 versus 0.573) and Economic status (0.633 versus 0.605) \rev{categories, while}
Population Activity is close to parity (0.863 versus 0.855)\rev{. Part of these gains may stem from overlap between model inputs and evaluation targets, as the CDC PLACES and economic targets are partly derived from census and ACS covariates that also inform the AGS inputs}.
At the target level, PlacePulse exceeds PDFM on three of the five Population Activity outcomes: Human Activity Index (0.935 versus 0.834), nighttime lights (0.955 versus 0.903), and population density (0.933 versus 0.865).
PDFM leads on raw population count (0.723 versus 0.825) and cell-tower density (0.768 versus 0.850), bringing the category-level average close to a tie.
Social capital is effectively tied across the two representations, while \rev{PDFM achieves higher $R^2$ on} Entrepreneurship (0.822 versus 0.765).

PDFM \rev{also achieves higher $R^2$} on Points of Interest (0.906 versus 0.784).
\rev{This pattern is consistent with a closer match between PDFM's inputs and these targets. PDFM's inputs include place counts from Google Maps \citep{agarwal2024}, which, like the Overture targets, are derived from listings of places. PlacePulse includes AGS establishment counts, which are derived from business records and classified by industry (NAICS) rather than by place category.}
Physical Geography also favors PDFM, especially for elevation, where it achieves $R^2 = 0.821$ compared with 0.555 for PlacePulse.
The $R^2$ spread across held-out folds is large in both cases, suggesting that ZCTA-level physical-geography prediction is difficult, mostly for PlacePulse, where little input contains information relative to this category.

The results suggest that the two embeddings behave as distinct model families rather than interchangeable representations.
PlacePulse, which is trained to encode structural relationships among place attributes, performs slightly better on Public Health, Economic status, and Population Activity.
These categories depend strongly on persistent socioeconomic, demographic, and institutional structure, and the gains are largest for health and economic outcomes.
PDFM performs better on Physical Geography, Points of Interest, and Entrepreneurship, with the largest advantage for POI counts.
This pattern is consistent with PDFM capturing more directly observable or dynamic signals of place intensity, land use, and economic activity, whereas PlacePulse captures more latent structural variation in the social and health environment.
\rev{These differences may also partly reflect the input data of each model, which in some cases overlap with the evaluation targets, either directly or through shared covariates (e.g., census-derived demographics for PlacePulse on health and economic targets, or map-derived POIs for PDFM and AGS establishment counts for PlacePulse on POI counts).}
Thus, the comparison should not be read simply as one embedding dominating the other; rather, the two model families appear to encode different but complementary aspects of place.

\begin{figure}[tbp]
\centering
\includegraphics[width=\linewidth]{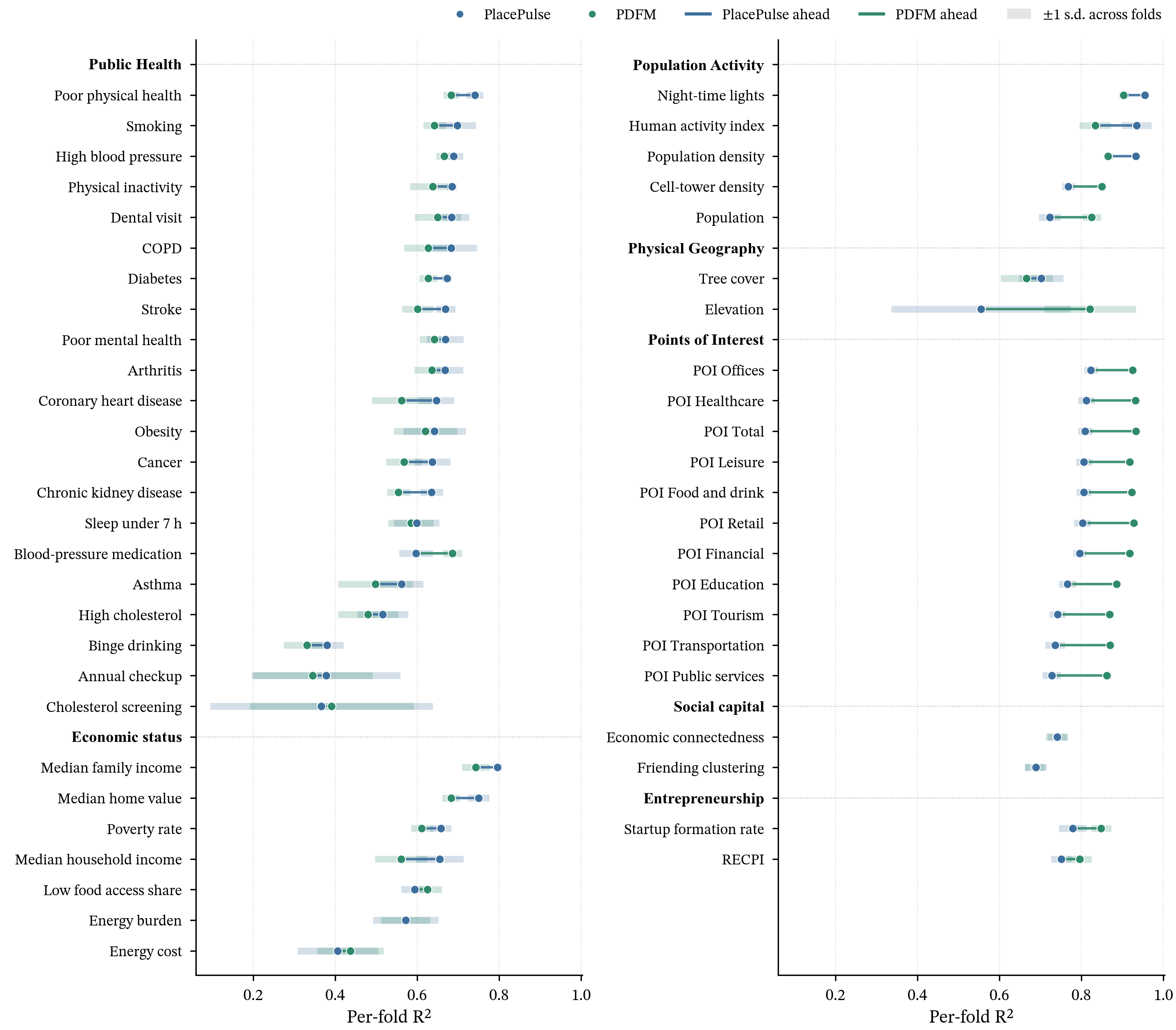}
\caption{\textbf{Downstream predictive performance at ZCTA resolution.}
PlacePulse embeddings, area-weighted from H3 resolution 7 to ZCTAs, are compared with Google PDFM embeddings at native ZCTA resolution.
The 50 targets are grouped into the seven ZCTA-level categories and split across two columns, under 5-fold spatially blocked cross-validation.
Each row is one target: the two dots are that target's per-fold mean $R^2$ under the two representations, the connector takes the color of the leading one, and the shaded band spans $\pm1$ standard deviation across the 5 held-out folds.}
\label{fig:down-zcta}
\end{figure}

\subsubsection{Other target resolutions (voting precinct, census tract)}
\label{sec:down-other}

For targets observed at voting-precinct and census-tract resolutions, neither representation is evaluated on its native spatial support.
\rev{Both PlacePulse and PDFM are transferred to the target units using the area-weighting procedure described in \xref{Section}{sec:eval}.}
The four targets span three categories, summarized in \xref{Table}{tab:down-other}.

\begin{table}[tbp]
\centering
\caption{\textbf{Downstream performance at non-native target resolutions: voting precincts and census tracts.}
Both representations are area-weighted from their original resolution (H3 res-7 for PlacePulse, ZCTA for PDFM) to the target resolution.
The average per-fold $R^2$ of the tasks in each category is reported; see \xref{Table}{tab:b3} for the uncertainty of each individual task.}
\label{tab:down-other}
\tabfont
\renewcommand{\arraystretch}{1.2}
\begin{tabularx}{\linewidth}{@{}Y{0.73} Y{1.27} r r r@{}}
\toprule
Category & What it covers & \# & PlacePulse $R^2$ & PDFM $R^2$ \\
\midrule
\textbf{Elections} \textit{(precinct)} & 2024 U.S. presidential result by voting precinct. & 1 & 0.737 & 0.709 \\
\textbf{Economic Mobility} \textit{(tract)} & Opportunity Atlas intergenerational mobility for children raised at the 25th income percentile. & 1 & 0.546 & 0.508 \\
\rev{\textbf{Social Vulnerability and Resilience}} \textit{(tract)} & Social-vulnerability and community-resilience scores. & 2 & 0.435 & 0.302 \\
\bottomrule
\end{tabularx}
\end{table}

PlacePulse outperforms PDFM in every category in the non-native-resolution evaluation.
For the 2024 presidential election result at the precinct level, PlacePulse achieves $R^2 = 0.737$, compared with 0.709 for PDFM, consistent with the embedding capturing structural sociodemographic variation strongly associated with spatial patterns in voting behavior.
For intergenerational economic mobility, PlacePulse also leads, with $R^2 = 0.546$ compared with 0.508 for PDFM.
\rev{Social Vulnerability and Resilience} shows the largest differences: PlacePulse \rev{leads on} both social vulnerability ($R^2 = 0.384$ versus 0.269) and community resilience ($R^2 = 0.486$ versus 0.335), while both models fail to estimate such targets correctly.
\rev{Both indices are built from census socioeconomic indicators that overlap with AGS inputs, which may contribute to the PlacePulse advantage.}

\rev{In this setting, both representations must be transferred from their original spatial supports to the target units.}
\rev{PlacePulse leads across all three categories and nearly all evaluation views, with the main exception being the shared failure of both representations to generalize on Social Vulnerability and Resilience.}
\rev{However, because each representation reaches the target units through a different transfer, these results are best read as a practical comparison: for analyses carried out at tract or precinct level, the pattern suggests that PlacePulse captures structural place information that is more stable across such spatial scales and target domains, particularly for outcomes shaped by persistent socioeconomic, demographic, and institutional context.}

\begin{figure}[tbp]
\centering
\includegraphics[width=0.675\linewidth]{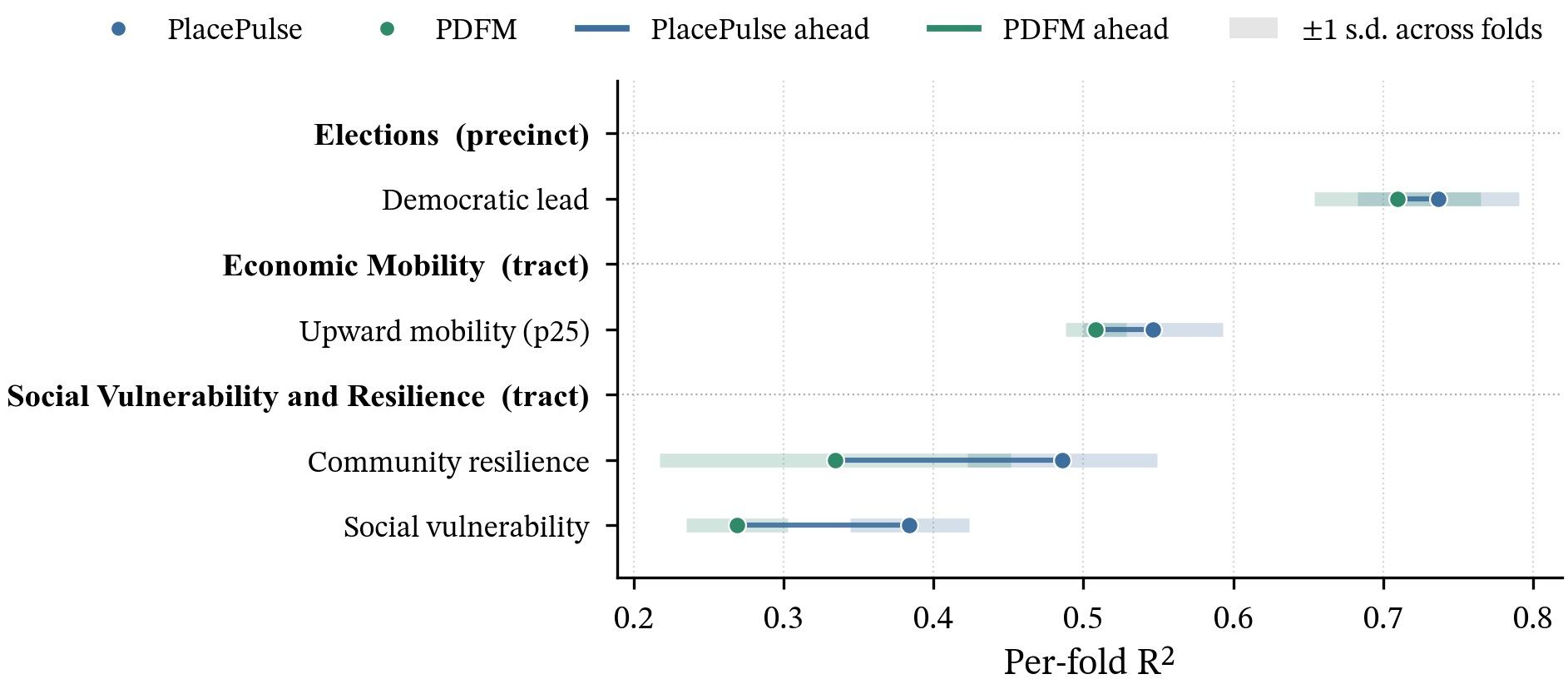}
\caption{\textbf{Downstream predictive performance at non-native target resolutions.}
PlacePulse and Google PDFM are both area-weighted from their native spatial supports to voting precincts or census tracts, under 5-fold spatially blocked cross-validation.
Each row is one target: the two dots are that target's per-fold mean $R^2$ under the two representations, the connector takes the color of the leading one, and the shaded band spans $\pm1$ standard deviation across the 5 held-out folds.}
\label{fig:down-other}
\end{figure}

\section{Discussion}
\label{sec:discussion}

\subsection{A compact, faithful re-encoding of structural place data}
\label{sec:disc-compact}

The internal evaluation shows that a purely structural autoencoder yields a faithful and well-conditioned representation of place.
The model reconstructs the 2,288 standardized inputs with an average $R^2$ of 0.853 (median 0.887), and no feature is reconstructed worse than its own mean, so the 256-dimensional bottleneck preserves the bulk of the structural signal rather than discarding it.
The resulting space is also well-behaved: it is unit-norm by construction, has no dead dimensions, and retains 251 of 256 mutually decorrelated axes.
Its effective dimensionality is nonetheless modest: the participation ratio rises from 6.6 in the input to 12.6 in the embedding while the nonlinear estimate falls from 32.3 to 25.2, indicating that the encoder redistributes a highly redundant input across a broader set of directions instead of piling variance onto a few.
The embedding is therefore a compact, near-lossless re-encoding of the structural feature set, which is the precondition for using it as a general-purpose feature.

The representation additionally recovers geographic organization that it was never explicitly trained to produce.
Global Moran's $I$ reaches 0.93 on the dominant principal component and exceeds 0.50 on every dimension but one, and a Getis--Ord $G_i^{*}$ analysis places 55.6\% of cells in significant clusters.
Crucially, this autocorrelation persists beyond the two-ring radius of the model's own inputs, so the smoothness reflects genuine broad-scale spatial organization inherited from the data (Tobler's first law) rather than an artefact of the ring-aggregate features.
K-means partitions of the space are correspondingly interpretable and balanced: they order locations along a settlement-density gradient, reach 94--98\% of maximum size entropy, and refine rather than reorganize as $k$ grows.
Settlement density is one recognizable axis of that partition, but it is not the only one: income separates suburban types that sit at comparable densities, so the embedding is not reducible to a density index.

\subsection{Structural and dynamic representations encode complementary aspects
of place}
\label{sec:disc-complementary}

These properties speak to the paper's central claim: that the structural composition of a place is a useful, self-contained modality for a geospatial foundation model.
Unlike the dynamic-signal and imagery models that dominate the field (\xref{Section}{sec:related}), PlacePulse requires no behavioral traces (which are sparse, privacy-sensitive, and unevenly available) and no imagery, yet it encodes who lives and works in a place together with its risk profile, information those modalities capture only indirectly.
The downstream comparison with PDFM supports reading the two as complementary rather than competing.
At ZCTA resolution PlacePulse leads on Public Health (0.610 versus 0.573) and Economic status (0.633 versus 0.605), is at parity on Population Activity (0.863 versus 0.855) and Social capital, and trails on Points of Interest (0.784 versus 0.906), Physical Geography (0.628 versus 0.744), and Entrepreneurship (0.765 versus 0.822).
The split is systematic: PlacePulse is stronger where outcomes depend on persistent socioeconomic, demographic, and institutional structure, and weaker where they depend on directly observable land use, physical form, or point-level activity intensity.

\rev{The spatial support of each comparison matters for its interpretation.}
\rev{At ZCTA, PlacePulse is evaluated after area-weighted aggregation from H3 resolution 7, whereas PDFM is scored on its native spatial support.}
\rev{At precinct and tract level, where both representations are transferred away from their native supports, PlacePulse leads in all three categories: 2024 presidential vote share at precinct level (0.737 versus 0.709), intergenerational economic mobility at tract level (0.546 versus 0.508), and social-vulnerability and resilience scores (0.435 versus 0.302).}
\rev{These results should be interpreted with the transfer in mind: PDFM reaches the target units from coarser ZCTAs that largely nest them, whereas PlacePulse reaches them from finer H3 cells that either cut across them or contain several of them within a single cell (\xref{Section}{sec:eval}), so the comparison reflects each representation together with its native support.}
\rev{The absolute performance on social vulnerability and resilience shows that both representations leave substantial variance unexplained.}
Testing whether a concatenated structural-plus-dynamic representation outperforms either alone is the natural next experiment, and we leave it to future work.
\rev{A second extension is to evaluate targets at coarser units, such as counties.}
\rev{Counties are larger than both native supports, so both representations would be aggregated rather than disaggregated, removing the boundary-misalignment and shared-embedding effects described above.}

\subsection{Implications for practice and responsible use}
\label{sec:disc-practice}

For practitioners, the appeal is a single reusable national representation that replaces the laborious, per-task feature engineering described in \xref{Section}{sec:intro}.
Rather than assembling, cleaning, and joining hundreds of variables for each new study, an analyst attaches a 256-dimensional vector to each H3 cell.
Because the embeddings lie on the unit sphere, cosine similarity reduces to a dot product, so retrieval tasks such as twin-area discovery run at national scale as a single matrix operation.
\rev{The neighborhood-coherence result also gives a practical licence for the aggregation step most applied work requires: area-weighted means over ZCTAs average coherent local vectors rather than unrelated ones, which supports the use of aggregated embeddings for ZCTA-level analyses such as those in \xref{Section}{sec:down-zcta}.}

Because the embeddings are learned by reconstructing sociodemographic inputs, they encode attributes, such as race and ethnicity, income, and a diversity index; and can act as proxies for them even when those attributes are never used explicitly downstream. This is a 
general property of representations learned from demographic data rather than a characteristic of this model in particular. The representation describes areas rather than individuals, although aggregation does not by itself remove the property. We intend the embeddings 
for spatial analysis, not for individual-level eligibility decisions in regulated domains, and those who deploy the embeddings are responsible for complying with applicable law, including any disparate-impact testing required in high-stakes settings, under CARTO's 
licensing terms. Treating the representation as attribute-free because no protected attribute appears in the model interface would be a serious misreading of what it contains.

\subsection{Limitations}
\label{sec:limitations}

Five limitations bound the scope of these claims.
First, the inputs are proprietary, modelled, single-vintage estimates: the AGS features are commercial estimates for 2025 rather than direct observations, so their own estimation error propagates into the embeddings and into every downstream result, and their licence limits full reproducibility.
We do not bound that input uncertainty here, and cross-validating core AGS variables against an independent source with published margins of error, such as the American Community Survey, remains outstanding.
Second, coverage is limited to the United States on a single national grid, so transferability to other countries and data regimes is untested.

Third, the scope of the evaluation is transductive: we produce embeddings for a fixed, fully observed grid and evaluate them as features over that grid.
The encoder can be applied to new inputs, but we do not validate its behaviour on unseen cells, vintages, or geographies, and we make no claim of inductive generalization.
Fourth, we report a single configuration selected from an exploratory sweep.
This follows from the scope set in \xref{Section}{sec:arch}, which treats the encoder as a conventional means to the representation rather than as a contribution in itself.
The sensitivity of the results to architecture and hyperparameters therefore remains only partially characterized. 
\rev{Fifth, some downstream targets share covariates with the inputs of PlacePulse, and in some cases PDFM (\xref{Table}{tab:targets}). Several were included deliberately, to check whether the embeddings preserve the socio-economic and health gradients present in the inputs (\xref{Section}{sec:targets}), but cross-model differences on these targets may partly reflect input-target overlap rather than representation quality alone; a controlled comparison excluding the overlapping inputs is left for future work.}

\section{Conclusions}
\label{sec:conclusion}

We introduced PlacePulse, a structural geospatial foundation model that compresses 572 AGS attributes (expanded to 2,288 inputs through H3 ring aggregates) into 256-dimensional, L2-normalized embeddings for 1,055,537 H3 resolution-7 cells in the United States.
The model uses no behavioral traces, imagery, or other dynamic inputs.
Intrinsically, the resulting embeddings provide a compact, well-conditioned, spatially coherent, and nearly lossless representation of the static composition of place.
Extrinsically, they serve as a competitive general-purpose feature representation across 74 downstream regression targets.
At the model's native resolution, PlacePulse outperforms a dimensionality-matched feature-selection baseline on targets unrelated to the training inputs.
\rev{After transfer to non-native spatial units, it matches or exceeds PDFM on health, economic, election, mobility, and social-vulnerability outcomes.}

The main contribution of this work is evidential rather than architectural.
Current geospatial foundation models are typically built around behavioral traces, mobility signals, or satellite imagery, while structural attributes are often treated as auxiliary covariates.
Our results show that structural composition alone can support a representation that is competitive with a dynamic-signal model on comparable spatial prediction tasks.
Together with a national-scale evaluation protocol that separates intrinsic representation quality from downstream predictive utility, these findings position structural embeddings as a distinct and complementary modality for geospatial foundation modeling, rather than merely as a substitute when behavioral or imagery data are unavailable.

Several directions follow from these results.
The most immediate is multimodal fusion.
Because structural and dynamic representations appear to capture complementary aspects of place, future work should directly test combinations of PlacePulse with PDFM-style, mobility-based, or Earth-observation embeddings; the category-level results reported here suggest where each modality is likely to contribute most.
\rev{A related direction is to characterize the contribution of each input family to the representation. Ablation studies that retrain the encoder with individual AGS products removed or used in isolation would clarify which thematic domains drive performance on each category of downstream task, help quantify the extent to which overlap between input covariates and evaluation targets contributes to that performance, and could guide the design of leaner or domain-specific variants of the model.} Another direction is temporal representation learning.
Extending PlacePulse to multiple data vintages would make it possible to learn temporally indexed embeddings of the same cells, separating persistent structural context from trajectories of change and enabling applications such as neighborhood change detection, early-warning indicators, and longitudinal comparison of places.
More explicitly, temporal encoders could also model sequences of structural states rather than treating each vintage independently.
Applying the approach in other countries would further test its portability across national data regimes.

On the modeling side, set-based or transformer-based encoders that omit masked inputs directly, rather than using the replace-in-place scheme adopted here, may improve both efficiency and representation quality.
A further direction concerns the geometry of the embedding space.
Although PlacePulse uses nearly all 256 dimensions, variance remains concentrated in a smaller number of directions, so the embeddings occupy only a narrow region of the unit sphere.
Future work could test uniformity- or entropy-based regularizers, such as the uniformity objective of \citet{wang2020uniformity} or the Kozachenko--Leonenko entropy regularizer of \citet{sablayrolles2019}, to spread representations while preserving local spatial structure.
However, greater uniformity is not necessarily preferable: the current density structure supports interpretable clustering, and some concentration likely reflects real structure in the input data.
Whether a more uniform embedding improves retrieval and downstream prediction, or instead removes meaningful variation, remains an open empirical question.

Finally, the unit-norm embedding geometry motivates dedicated retrieval applications, such as twin-area discovery and market-whitespace analysis, as well as systematic evaluation of methods for aggregating cell-level embeddings to policy-relevant spatial units and quantifying how much fine-scale structure is lost in that process.

\section*{Acknowledgements}

We thank Applied Geographic Solutions for granting access to the data products described in \xref{Section}{sec:data}, and for reviewing our description of those products.
We thank Esteban Moro and Hamish Gibbs (Network Science Institute, Northeastern University) for insightful discussions on this work. 
We also thank Alex Singleton (University of Liverpool) for suggesting downstream evaluation targets, and Bruno Sánchez-Andrade (LGND) for discussions on the modeling approach.

We further thank Google Research for the public release of the Population Dynamics Foundation Model embeddings and benchmark targets, and the providers of the open data used as 
downstream evaluation targets (\xref{Section}{sec:targets}): the Overture Maps Foundation; OpenCellID; the Earth Observation Group at the Payne Institute for Public Policy, Colorado School of 
Mines; the U.S. Department of Agriculture Economic Research Service; the National Renewable Energy Laboratory and the U.S. Department of Energy; The New York Times; Opportunity Insights 
(Opportunity Atlas and Social Capital Atlas); the Federal Emergency Management Agency; and the Startup Cartography Project.

\section*{Declaration of competing interest}

The authors declare the following financial interests and personal relationships which may be considered as potential competing interests.
All authors are employees of CARTO, which developed the PlacePulse embeddings reported in this paper and holds a commercial interest in their distribution as a data product.
The input features are licensed from Applied Geographic Solutions, a commercial data provider with which CARTO has a commercial partnership for the distribution of Applied Geographic Solutions data products. 
Applied Geographic Solutions reviewed the description of its own data products in \xref{Section}{sec:data} and had no role in the study design, the evaluation protocol, the analysis or 
interpretation of the results, or the decision to submit this article for publication.

\section*{Funding}

This work was funded by CARTO as part of its internal research programme. We acknowledge the EuroHPC Joint Undertaking for awarding two projects (EHPC-AIF-2025PG01-473 and EHPC-AIF-2026PG01-587) access to the EuroHPC supercomputer MareNostrum 5, hosted by the Barcelona Supercomputing Center (Spain), through the EuroHPC AI Factory Playground Access mode.

Apart from this computing allocation, the work received no specific grant from external funding agencies. CARTO was involved in the study design, the analysis and interpretation of the results, the writing of the article, and the decision to submit it for publication; the EuroHPC Joint Undertaking had no role in any of these.

\section*{Data availability}

The input features are licensed from Applied Geographic Solutions and cannot be redistributed by the authors; access requires a licence agreement with the provider.
The trained embeddings are a CARTO data product and are not publicly redistributable, and the code that implements the model is not publicly available.
The downstream evaluation targets are drawn from the public sources listed in \xref{Table}{tab:targets}, with the exception of the six AGS Health targets, and each remains available from its original provider under that provider's terms.
The Population Dynamics Foundation Model embeddings used for comparison are distributed by Google Research under their own access terms.

\section*{Declaration of generative AI and AI-assisted technologies in the
manuscript preparation process}

During the preparation of this work the authors used ChatGPT (OpenAI) and Claude (Anthropic) in order to enhance the readability and language of the manuscript.
After using these tools, the authors reviewed and edited the content as needed and take full responsibility for the content of the published article.

\section*{Author Contributions (CRediT)}

\textbf{Luc\'{\i}a Garc\'{\i}a-Duarte:} Conceptualization, Data curation, Formal analysis, Investigation, Methodology, Software, Validation, Visualization, Writing - original draft, Writing - review \& editing.
\textbf{Miguel \'Alvarez:} Conceptualization, Funding acquisition, Project administration, Supervision, Writing - original draft, Writing - review \& editing.
\textbf{Diego Vicente:} Conceptualization, Data curation, Formal analysis, Writing - review \& editing. 
\textbf{Javier P\'erez Trufero:} Conceptualization, Supervision.
\textbf{Javier de la Torre:} Conceptualization, Funding acquisition, Supervision.

\clearpage
\appendix
\renewcommand{\thefigure}{\thesection.\arabic{figure}}
\renewcommand{\thetable}{\thesection.\arabic{table}}
\setcounter{figure}{0}
\setcounter{table}{0}

\section{Supplemental clustering structure evaluation}
\label{app:clustering}

\xref{Section}{sec:spatial} reports that finer K-means partitions refine coarser ones rather than reorganizing them.
This appendix presents the evidence behind that statement, for the three transitions in granularity $k = 10 \rightarrow 50$, $k = 10 \rightarrow 100$, and $k = 50 \rightarrow 100$.

The nesting matrices are strongly block-diagonal (\xref[a]{Figure}{fig:nesting}): most fine clusters draw the majority of their cells from a single coarse cluster, and mixing across parents is limited in all three scenarios.

Purity quantifies the pattern: the empirical cumulative distribution in \xref[b]{Figure}{fig:nesting} shows fine-cluster purity, defined as the largest share of a fine cluster's cells coming from a single coarse cluster.
Higher purity means that a fine cluster is mostly nested within one coarse parent, rather than being assembled from multiple coarse clusters.
Median purity is 68.4\% for $k = 10 \rightarrow 50$, 70.8\% for $k = 10 \rightarrow 100$, and 54.6\% for $k = 50 \rightarrow 100$, indicating that a typical fine cluster draws the majority of its cells from one parent partition.
The chance baseline is the cell share of the largest cluster in the source partition, 18.5\% for $k = 10$ and 3.7\% for $k = 50$, so the observed medians exceed chance by $3.7\times$, $3.8\times$, and $14.7\times$, showing that the nesting pattern is far stronger than would be expected from random overlap.
Because purity depends on the relative granularities of the two partitions, the three medians should be interpreted against their own baselines rather than compared directly.
In particular, $k = 50 \rightarrow 100$ has the lowest absolute median purity but is furthest above chance, indicating that even the finest transition retains substantial hierarchical structure.
Overall, the purity distributions confirm that finer K-means solutions are not arbitrary reorganizations of the data; they largely preserve the structure of coarser partitions while introducing additional subdivision.

\xref[c]{Figure}{fig:nesting} shows how larger coarse clusters tend to produce more fine-cluster descendants.
The relationship between coarse-cluster size and number of children is positive for all transitions, with Pearson correlations of 0.85 for $k = 10 \rightarrow 50$, 0.88 for $k = 10 \rightarrow 100$, and 0.49 for $k = 50 \rightarrow 100$.
This indicates that larger coarse clusters are generally subdivided into more fine clusters, while smaller coarse clusters tend to remain represented by fewer descendants.
The weaker correlation for $k = 50 \rightarrow 100$ suggests that, at finer granularities, splitting is less determined by size alone and may reflect additional internal heterogeneity within the coarse clusters.
Overall, the pattern supports the interpretation that increasing $k$ primarily refines existing clusters, with larger parent clusters receiving more subdivisions rather than the partition being globally reorganized.

Taken together, the three views show that K-means solutions at increasing $k$ are not arbitrary reorganizations of the data.
They preserve the structure of coarser partitions and add subdivision, even though K-means imposes no hierarchical constraint.

\begin{figure}[tbp]
\centering
\includegraphics[width=\linewidth]{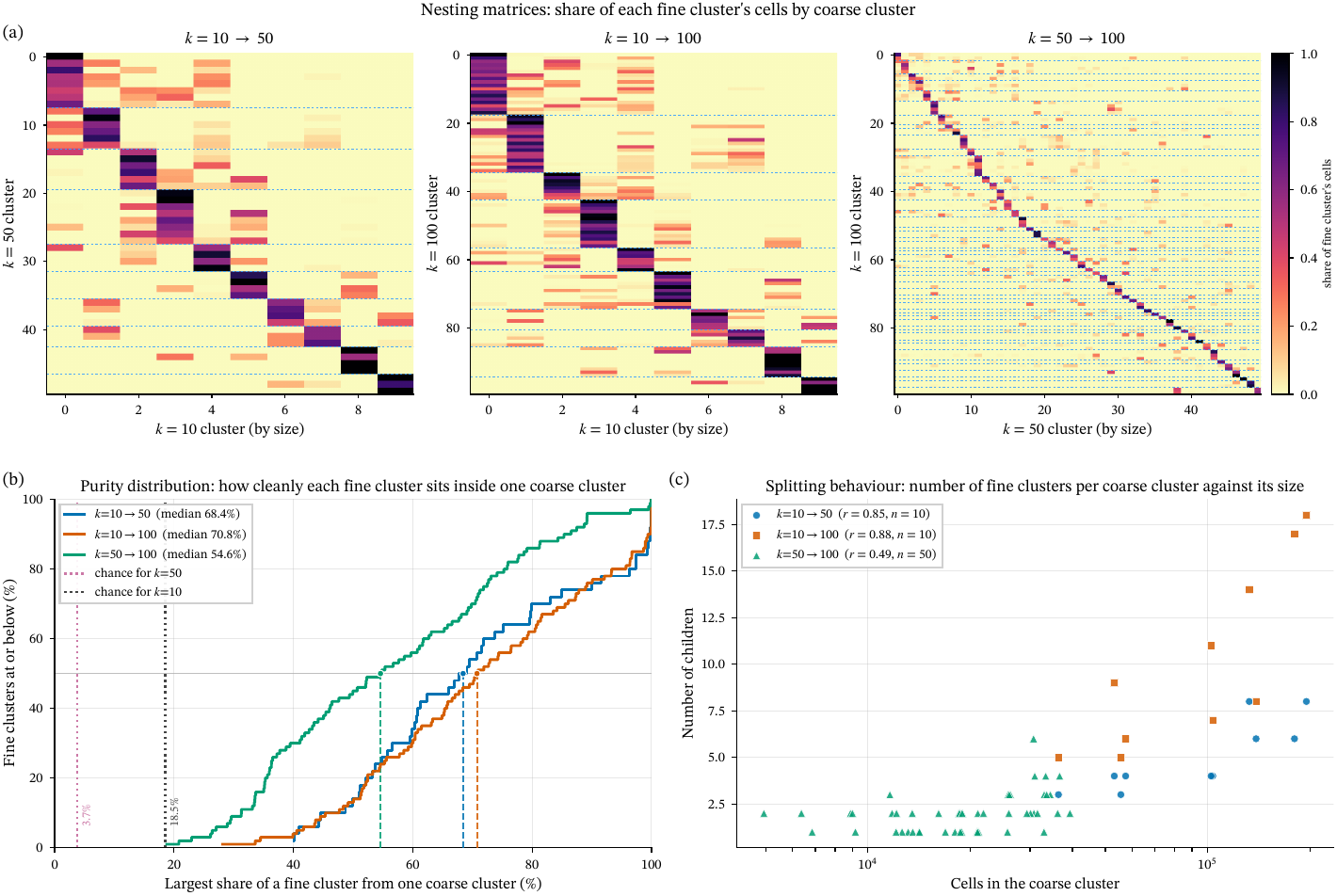}
\caption{\textbf{Finer K-means partitions mostly refine coarser partitions rather than reorganizing them.}
We compare three transitions in clustering granularity: $k = 10 \rightarrow 50$, $k = 10 \rightarrow 100$, and $k = 50 \rightarrow 100$.
(a) Nesting matrices.
Rows represent fine clusters and columns represent coarse clusters, ordered by coarse-cluster size.
Shading indicates the share of each fine cluster's cells that fall within each coarse cluster.
Rows are sorted by their dominant coarse ``parent'', and dashed lines separate the resulting parent blocks.
(b) Empirical cumulative distribution of fine-cluster purity, defined as the largest share of a fine cluster's cells coming from a single coarse cluster.
Dashed vertical lines mark the median of each transition, and dotted vertical lines mark the chance baseline, the cell share of the largest cluster in the source partition.
(c) Number of fine-cluster descendants against coarse-cluster size, on a logarithmic x-axis.
Marker shape and color identify the transition, and the legend gives the Pearson correlation $r$ and the number of coarse clusters $n$.}
\label{fig:nesting}
\end{figure}

\section{Supplemental downstream predictive evaluation}
\label{app:downstream}

\xref{Section}{sec:downstream} reports downstream performance aggregated by category.
This appendix gives the result for every individual target, so a reader can see which targets drive each category mean.
\xref{Table}{tab:b1} covers the 20 targets at the model's native H3 resolution 7, \xref{Table}{tab:b2} the 50 targets at ZCTA resolution, and \xref{Table}{tab:b3} the 4 targets at voting-precinct and census-tract resolution.

Every value is the per-fold $R^2$, the mean $\pm$ standard deviation of $R^2$ computed within each held-out fold of the 5-fold spatially-blocked cross-validation, where every state is held out entirely in one fold.
This is the same view \xref{Section}{sec:downstream} reports at category level.
Both representations are always scored on the same GEOIDs, the intersection of the units each one covers, so neither is ed with an easier test set.

{\tabfont
\renewcommand{\arraystretch}{1.2}
\begin{xltabular}{\linewidth}{@{}Y{0.72} Y{1.13} Y{0.95} Y{1.20} r r@{}}
\caption{\textbf{Per-target downstream performance at native H3 resolution~7.}
PlacePulse (256-d) against the target-blind, dimensionality-matched baseline, under 5-fold spatially-blocked cross-validation with every state held out entirely in one fold. Points of Interest targets are scored in log space. In every row the higher per-fold mean is set in bold; the one row where the two means agree to three decimals is left unmarked.}\label{tab:b1}\\
\midrule
Category & Source & Target & Description & PlacePulse $R^2$ & Baseline $R^2$ \\
\midrule
\endfirsthead
\toprule
Category & Source & Target & Description & PlacePulse $R^2$ & Baseline $R^2$ \\
\midrule
\endhead
\toprule
\endfoot
\bottomrule
\endlastfoot
\textbf{Public Health} & Applied Geographic Solutions & High blood pressure (risk) & Adults with high blood pressure & 0.729 $\pm$ 0.020 & \textbf{0.759} $\pm$ 0.020 \\
 &  & Cancer (risk) & Adults with cancer, excluding skin cancer & 0.712 $\pm$ 0.023 & \textbf{0.741} $\pm$ 0.021 \\
 &  & High cholesterol (risk) & Adults with high cholesterol & 0.767 $\pm$ 0.015 & \textbf{0.791} $\pm$ 0.012 \\
 &  & Diabetes (risk) & Adults with diagnosed diabetes & 0.670 $\pm$ 0.036 & \textbf{0.710} $\pm$ 0.028 \\
 &  & Obesity (risk) & Adults classified obese & 0.705 $\pm$ 0.018 & \textbf{0.733} $\pm$ 0.006 \\
 &  & Stroke (risk) & Adults who have had a stroke & 0.661 $\pm$ 0.018 & \textbf{0.695} $\pm$ 0.015 \\
\textbf{Points of Interest} & Overture Maps & POI Total & All POI categories combined & \textbf{0.820} $\pm$ 0.016 & 0.790 $\pm$ 0.017 \\
 &  & POI Retail & Shops and retail outlets & \textbf{0.803} $\pm$ 0.018 & 0.757 $\pm$ 0.019 \\
 &  & POI Healthcare & Clinics, hospitals, pharmacies & \textbf{0.823} $\pm$ 0.017 & 0.760 $\pm$ 0.018 \\
 &  & POI Offices & Office and professional-services premises & \textbf{0.833} $\pm$ 0.015 & 0.776 $\pm$ 0.017 \\
 &  & POI Food and drink & Restaurants, cafes, bars & \textbf{0.797} $\pm$ 0.023 & 0.737 $\pm$ 0.025 \\
 &  & POI Financial & Banks, ATMs, financial services & \textbf{0.790} $\pm$ 0.022 & 0.716 $\pm$ 0.023 \\
 &  & POI Leisure & Sports, recreation, entertainment & \textbf{0.779} $\pm$ 0.025 & 0.740 $\pm$ 0.025 \\
 &  & POI Education & Schools, universities, training & \textbf{0.762} $\pm$ 0.028 & 0.716 $\pm$ 0.027 \\
 &  & POI Public services & Government and civic facilities & \textbf{0.727} $\pm$ 0.022 & 0.678 $\pm$ 0.021 \\
 &  & POI Tourism & Hotels, attractions, tourist services & \textbf{0.689} $\pm$ 0.037 & 0.667 $\pm$ 0.034 \\
 &  & POI Transportation & Stations, stops, transport hubs & \textbf{0.671} $\pm$ 0.035 & 0.622 $\pm$ 0.036 \\
\textbf{Population Activity} & CARTO Spatial Features & Human activity index & Composite human-presence proxy & \textbf{0.930} $\pm$ 0.009 & 0.911 $\pm$ 0.007 \\
 &  & Nighttime lights & Nighttime lights: VIIRS average radiance & \textbf{0.769} $\pm$ 0.059 & 0.758 $\pm$ 0.050 \\
 &  & Cell-tower density & Cell-tower density & \textbf{0.704} $\pm$ 0.024 & 0.696 $\pm$ 0.023 \\
\end{xltabular}
}

{\tabfont
\renewcommand{\arraystretch}{1.2}
\begin{xltabular}{\linewidth}{@{}Y{0.72} Y{1.13} Y{0.95} Y{1.20} r r@{}}
\caption{\textbf{Per-target downstream performance at ZCTA resolution.}
PlacePulse embeddings are area-weighted from H3 resolution~7 to ZCTAs; PDFM embeddings are used at their native ZCTA resolution. Reporting conventions match \xref{Table}{tab:b1}. Population Activity, Points of Interest, and Entrepreneurship targets are scored in log space.}\label{tab:b2}\\
\midrule
Category & Source & Target & Description & PlacePulse $R^2$ & PDFM $R^2$ \\
\midrule
\endfirsthead
\toprule
Category & Source & Target & Description & PlacePulse $R^2$ & PDFM $R^2$ \\
\midrule
\endhead
\toprule
\endfoot
\bottomrule
\endlastfoot
\textbf{Public Health} & PDFM benchmark \texttt{conus27} & Poor physical health & Adults reporting poor physical health & \textbf{0.741} $\pm$ 0.021 & 0.683 $\pm$ 0.019 \\
 &  & Poor mental health & Adults reporting poor mental health & \textbf{0.669} $\pm$ 0.045 & 0.642 $\pm$ 0.036 \\
 &  & High blood pressure & Adults with high blood pressure & \textbf{0.689} $\pm$ 0.023 & 0.666 $\pm$ 0.021 \\
 &  & High cholesterol & Adults with high cholesterol & \textbf{0.516} $\pm$ 0.062 & 0.481 $\pm$ 0.074 \\
 &  & Diabetes & Adults with diagnosed diabetes & \textbf{0.673} $\pm$ 0.012 & 0.627 $\pm$ 0.022 \\
 &  & Obesity & Adults classified obese & \textbf{0.642} $\pm$ 0.076 & 0.621 $\pm$ 0.078 \\
 &  & Cancer & Adults with cancer, excluding skin cancer & \textbf{0.638} $\pm$ 0.043 & 0.568 $\pm$ 0.044 \\
 &  & Stroke & Adults who have had a stroke & \textbf{0.669} $\pm$ 0.023 & 0.601 $\pm$ 0.038 \\
 &  & Coronary heart disease & Adults with coronary heart disease & \textbf{0.647} $\pm$ 0.044 & 0.562 $\pm$ 0.073 \\
 &  & Chronic kidney disease & Adults with chronic kidney disease & \textbf{0.635} $\pm$ 0.027 & 0.554 $\pm$ 0.029 \\
 &  & COPD & Adults with COPD & \textbf{0.683} $\pm$ 0.063 & 0.627 $\pm$ 0.059 \\
 &  & Asthma & Adults with current asthma & \textbf{0.562} $\pm$ 0.053 & 0.498 $\pm$ 0.091 \\
 &  & Arthritis & Adults with arthritis & \textbf{0.668} $\pm$ 0.043 & 0.636 $\pm$ 0.043 \\
 &  & Smoking & Adults who currently smoke & \textbf{0.698} $\pm$ 0.045 & 0.642 $\pm$ 0.027 \\
 &  & Binge drinking & Adults reporting binge drinking & \textbf{0.381} $\pm$ 0.039 & 0.332 $\pm$ 0.058 \\
 &  & Physical inactivity & Adults reporting no leisure-time physical activity & \textbf{0.685} $\pm$ 0.013 & 0.638 $\pm$ 0.056 \\
 &  & Sleep under 7 h & Adults sleeping under 7 hours & \textbf{0.599} $\pm$ 0.055 & 0.585 $\pm$ 0.056 \\
 &  & Annual checkup & Adults who had a routine checkup & \textbf{0.378} $\pm$ 0.181 & 0.346 $\pm$ 0.146 \\
 &  & Dental visit & Adults who visited a dentist & \textbf{0.684} $\pm$ 0.043 & 0.651 $\pm$ 0.057 \\
 &  & Cholesterol screening & Adults screened for cholesterol & 0.367 $\pm$ 0.271 & \textbf{0.392} $\pm$ 0.200 \\
 &  & Blood-pressure medication & Hypertensive adults taking BP medication & 0.597 $\pm$ 0.041 & \textbf{0.686} $\pm$ 0.023 \\
\textbf{Economic status} & USDA ERS Food Access Research Atlas (2019) & Median family income & Median family income, ACS-derived (USD) & \textbf{0.796} $\pm$ 0.011 & 0.743 $\pm$ 0.034 \\
 &  & Low food access share & Share of population far from a supermarket ($>$1 mi urban / $>$10 mi rural) & 0.594 $\pm$ 0.032 & \textbf{0.625} $\pm$ 0.035 \\
 &  & Poverty rate & Share of population below the federal poverty line & \textbf{0.658} $\pm$ 0.025 & 0.611 $\pm$ 0.026 \\
 & PDFM benchmark \texttt{conus27} & Median household income & Median household income (USD) & \textbf{0.656} $\pm$ 0.057 & 0.561 $\pm$ 0.064 \\
 &  & Median home value & Median value of owner-occupied homes (USD) & \textbf{0.750} $\pm$ 0.026 & 0.683 $\pm$ 0.021 \\
 & NREL / U.S. DOE LEAD Tool (2022) & Energy burden & Average household energy cost as a share of income & 0.572 $\pm$ 0.080 & 0.572 $\pm$ 0.059 \\
 &  & Energy cost & Average annual household energy cost (USD) & 0.407 $\pm$ 0.099 & \textbf{0.438} $\pm$ 0.081 \\
\textbf{Population Activity} & PDFM benchmark \texttt{conus27} & Population & Resident population count & 0.723 $\pm$ 0.027 & \textbf{0.825} $\pm$ 0.023 \\
 &  & Population density & Population density, log10 & \textbf{0.933} $\pm$ 0.008 & 0.865 $\pm$ 0.011 \\
 &  & Nighttime lights & Nighttime lights radiance, log10 & \textbf{0.955} $\pm$ 0.004 & 0.903 $\pm$ 0.012 \\
 & CARTO Spatial Features & Human activity index & Human Activity Index: composite human-presence proxy & \textbf{0.935} $\pm$ 0.037 & 0.834 $\pm$ 0.038 \\
 &  & Cell-tower density & Cell-tower density in the unit & 0.768 $\pm$ 0.016 & \textbf{0.850} $\pm$ 0.007 \\
\textbf{Physical Geography} & PDFM benchmark \texttt{conus27} & Elevation & Mean ground elevation (m) & 0.555 $\pm$ 0.218 & \textbf{0.821} $\pm$ 0.112 \\
 &  & Tree cover & Share of area under tree canopy & \textbf{0.702} $\pm$ 0.054 & 0.667 $\pm$ 0.064 \\
\textbf{Points of Interest} & Overture Maps & POI Total & All POI categories combined & 0.810 $\pm$ 0.017 & \textbf{0.933} $\pm$ 0.005 \\
 &  & POI Retail & Shops and retail outlets & 0.803 $\pm$ 0.020 & \textbf{0.928} $\pm$ 0.005 \\
 &  & POI Healthcare & Clinics, hospitals, pharmacies & 0.812 $\pm$ 0.020 & \textbf{0.932} $\pm$ 0.005 \\
 &  & POI Offices & Office and professional-services premises & 0.823 $\pm$ 0.016 & \textbf{0.925} $\pm$ 0.005 \\
 &  & POI Food and drink & Restaurants, cafes, bars & 0.806 $\pm$ 0.018 & \textbf{0.923} $\pm$ 0.004 \\
 &  & POI Financial & Banks, ATMs, financial services & 0.796 $\pm$ 0.016 & \textbf{0.919} $\pm$ 0.006 \\
 &  & POI Leisure & Sports, recreation, entertainment & 0.806 $\pm$ 0.019 & \textbf{0.918} $\pm$ 0.004 \\
 &  & POI Education & Schools, universities, training & 0.766 $\pm$ 0.021 & \textbf{0.886} $\pm$ 0.007 \\
 &  & POI Public services & Government and civic facilities & 0.728 $\pm$ 0.022 & \textbf{0.862} $\pm$ 0.004 \\
 &  & POI Tourism & Hotels, attractions, tourist services & 0.742 $\pm$ 0.019 & \textbf{0.869} $\pm$ 0.010 \\
 &  & POI Transportation & Stations, stops, transport hubs & 0.736 $\pm$ 0.024 & \textbf{0.870} $\pm$ 0.004 \\
\textbf{Social capital} & Social Capital Atlas & Economic connectedness & Share of high-socioeconomic-status friends among low-status residents & \textbf{0.741} $\pm$ 0.024 & 0.740 $\pm$ 0.027 \\
 &  & Friending clustering & Friendship clustering coefficient: how interconnected a resident's friends are & \textbf{0.689} $\pm$ 0.027 & 0.686 $\pm$ 0.025 \\
\textbf{Entrepre-neurship} & Startup Cartography Project & Startup formation rate & Startup Formation Rate: new business registrations (\texttt{log1p}) & 0.779 $\pm$ 0.034 & \textbf{0.848} $\pm$ 0.025 \\
 &  & RECPI & Regional Entrepreneurship Cohort Potential Index (log) & 0.751 $\pm$ 0.025 & \textbf{0.796} $\pm$ 0.030 \\
\end{xltabular}
}

{\tabfont
\renewcommand{\arraystretch}{1.2}
\begin{xltabular}{\linewidth}{@{}Y{0.72} Y{1.13} Y{0.95} Y{1.20} r r@{}}
\caption{\textbf{Per-target downstream performance at non-native target resolutions.}
Both PlacePulse and PDFM are area-weighted from their native spatial supports to voting precincts or census tracts. Reporting conventions match \xref{Table}{tab:b1}.}\label{tab:b3}\\
\midrule
Category & Source & Target & Description & PlacePulse $R^2$ & PDFM $R^2$ \\
\midrule
\endfirsthead
\toprule
Category & Source & Target & Description & PlacePulse $R^2$ & PDFM $R^2$ \\
\midrule
\endhead
\toprule
\endfoot
\bottomrule
\endlastfoot
\textbf{Elections} & New York Times 2024 presidential precinct map & Democratic lead & Democratic minus Republican vote share, 2024 presidential & \textbf{0.737} $\pm$ 0.054 & 0.709 $\pm$ 0.055 \\
\textbf{Economic Mobility} & Opportunity Atlas & Upward mobility (p25) & Mean adult income rank for children raised at the 25th income percentile & \textbf{0.546} $\pm$ 0.046 & 0.508 $\pm$ 0.020 \\
\rev{\textbf{Social Vulnerability and Resilience}} & FEMA National Risk Index & Social vulnerability & FEMA NRI social vulnerability score & \textbf{0.384} $\pm$ 0.040 & 0.269 $\pm$ 0.034 \\
 &  & Community resilience & FEMA NRI community resilience score & \textbf{0.486} $\pm$ 0.063 & 0.335 $\pm$ 0.117 \\
\end{xltabular}
}


\clearpage
\bibliographystyle{plainnat}
\bibliography{refs}

\end{document}